\documentclass{article}
\usepackage{amsmath}
 \usepackage[preprint]{neurips_2026}

\usepackage[utf8]{inputenc} % allow utf-8 input
\usepackage[T1]{fontenc}    % use 8-bit T1 fonts
\usepackage{hyperref}       % hyperlinks
\usepackage{url}            % simple URL typesetting
\usepackage{booktabs}       % professional-quality tables
\usepackage{amsfonts}       % blackboard math symbols
\usepackage{nicefrac}       % compact symbols for 1/2, etc.
\usepackage{microtype}      % microtypography
\usepackage{xcolor}         % colors
\usepackage{graphicx}
\usepackage{subcaption}
\usepackage{booktabs}
\usepackage[table]{xcolor}
\usepackage{multirow}
\usepackage{comment}
\usepackage{nicematrix}
\usepackage{tikz}
\usetikzlibrary{calc}

\title{High-Fidelity Surface Splatting-Based 3D Reconstruction from Multi-View Images}

\author{%
   Nandhana Sunil\thanks{Equal Contribution} \\
   IIT Palakkad, India \\
   \texttt{112201008@smail.iitpkd.ac.in} \\
   \And
   Abhirami R Iyer\footnotemark[1] \\
   IIT Palakkad, India \\
   \texttt{112201001@smail.iitpkd.ac.in} \\
   \And
   Avirup Mandal \\
   IIT Palakkad, India \\
   \texttt{mandal.avirup@gmail.com} \\
}
\begin{document}

\maketitle

\begin{abstract}
Multi-view mesh reconstruction remains a core challenge in computer graphics and vision, especially for recovering high-frequency geometry from sparse observations. Recent methods such as 3D Gaussian Splatting (3DGS) and Neural Radiance Fields (NeRF) rely on post-processing for mesh extraction, thereby limiting joint optimization of geometry and appearance. Implicit Moving Least Squares (IMLS) instead enables direct conversion of point clouds into signed distance and texture fields, supporting end-to-end reconstruction and rendering. However, existing IMLS formulations use exponential kernels that struggle with high-frequency detail. We introduce a compact polynomial kernel with local support and greater flexibility, allowing better control over frequency content and improved geometric fidelity. To further enhance fine details, we incorporate stochastic regularization with Laplacian filtering. Together, these improve the preservation of high-frequency structure while maintaining stable optimization. Experiments show state-of-the-art performance in both surface reconstruction and rendering, yielding more accurate geometry and sharper visuals from multi-view data.

%{\begin{center}
    %\textcolor{blue}{\href{https://anonymous.4open.science/r/Improved-IMLS-Splatting-92EB/README.md}{anonymous.4open.science/r/Improved-IMLS-Splatting-92EB/}}
    %\textcolor{blue}{\href{https://anonymous.4open.science/r/Improved-IMLS-Splatting-C541/README.md}{anonymous.4open.science/r/Improved-IMLS-C541/}}
%\end{center}} 
\end{abstract}

\section{Introduction}\label{sec:intro}

Reconstructing accurate 3D surfaces from multi-view images is a fundamental problem in computer vision and graphics, laying the foundation for applications ranging from scene interpretation to photorealistic rendering. Recent advances in differentiable rendering have enabled end-to-end mesh optimization pipelines that directly supervise surface geometry through image reconstruction losses. Among those differentiable methods, IMLS Splatting~\cite{yang2025imls} is an efficient approach that bridges the gap between sparse point clouds and continuous surfaces using the Implicit Moving Least Squares (IMLS) formulation, achieving state-of-the-art reconstruction quality.

An integral part of the IMLS formulation is the kernel function that determines each point's spatially weighted contribution. IMLS splatting employees an exponential kernel, which is mathematically convenient but has certain well-known limitations. First, the exponential kernel has infinite support, as its influence theoretically extends across the entire spatial domain. This could introduce structural artifacts and gradient discontinuities at the boundaries where the kernel's support radius gets arbitrarily truncated. Second, exponential kernels cannot dynamically adapt to local geometric complexity, limiting the algorithm's ability to capture high-frequency surface details and fine-grained structural variations. So, the optimization landscape would be particularly susceptible to local minima, as it involves a truncation boundary.

In this work, we address these limitations by replacing the exponential kernel with a novel compact polynomial kernel that offers key mathematical advantages. The proposed kernel~\cite{cheng2014kernel} has inherent finite support, adaptive parameterization, and additional degrees of freedom. By introducing per-point learnable shape parameters, our kernel can dynamically adapt its decay profile to local surface geometry, enabling sharper, more localized weighting than the exponential kernel.

%cite properly
Now that the kernel has compact support, we have to mitigate the challenges posed by a landscape susceptible to local minima. To address this, we adopt the stochastic preconditioning technique by~\citet{ling2025stoc}, where, during training, each voxel is perturbed by a small uniform noise. This smoothens the loss landscape early in training and ensures that the objective is minimized to the desired value. On the other hand, high-frequency details, such as sharp, thin edges, may be suppressed due to the smoothness introduced by stochastic preconditioning. To recover, we apply a Laplacian filtering to the implicit field to preserve the sharp features on the loss manifold.

We summarize our contribution as follows:
\begin{itemize}
    \item We propose a novel compact polynomial kernel with inherent finite support and adaptive per-point parameterization, overcoming the core limitations of the exponential kernel used in IMLS-Splatting.
    \item We integrate stochastic preconditioning into the neural framework, smoothing the optimization landscape and helping the field resolve low-frequency structure before fine detail.
    \item We incorporate Laplacian filtering as a residual correction to recover fine geometric detail.
    \item We evaluate our method on the NeRF synthetic dataset~\cite{mildenhall2020nerfrepresentingscenesneural} and the DTU dataset~\cite{jensen2014large} and demonstrate that our novel kernel produces meshes with finer surface detail and fewer artifacts than IMLS-Splatting.
\end{itemize}

\section{Related Work}\label{sec:relwork}

Implicit Moving Least-Squares~\cite{kolluri2008mls} is a classical point-set surface method that constructs a global signed distance field by combining the signed projection distances of tangent planes at each point, weighted by a kernel function. Theoretically, it guarantees smoothness and continuity, making surface reconstruction more feasible. Subsequent works such as Deep IMLS~\cite{liu2021deepimls} and Neural-IMLS~\cite{wang2021imls} incorporate IMLS into neural reconstruction pipelines, using ground-truth signed distance fields (SDF) to supervise the field and achieve high-quality geometric reconstruction. However, these methods rely on ground-truth surface supervision and do not support end-to-end optimization from multi-view images. IMLS-Splatting~\cite{yang2025imls} addresses this by introducing a fast splatting-based IMLS algorithm that enables differentiable mesh reconstruction directly from multi-view images. Our work builds directly on IMLS-Splatting, replacing its kernel with a more expressive kernel derived from the compactly supported weight function of \cite{cheng2014kernel}, which was itself inspired by Wendland radial basis functions~\cite{Wendland1995}. 

Neural view synthesis methods have achieved impressive results in multi-view reconstruction by representing scenes as continuous implicit functions optimized through volumetric rendering. NeRF~\cite{mildenhall2020nerfrepresentingscenesneural} models scenes as volumetric radiance fields, enabling high-quality novel view synthesis but struggling to produce well-defined surfaces. SDF-based methods, such as NeuS~\cite{wang2015pd}~\cite{yariv2021volume}, address this by reparameterizing the density field as a signed distance function, thereby enabling more accurate surface extraction. Neuralangelo~\cite{li2023neuralangelohighfidelityneuralsurface} further improves surface quality by incorporating multi-resolution hash grids and high-order numerical gradients. While these methods produce high-quality surfaces, they rely on dense volumetric sampling, leading to high computational costs and slow training times. Furthermore, meshes are extracted through a post-processing step independent of the optimization, potentially introducing additional geometric errors.

3DGS~\cite{kerbl20233dgaussiansplattingrealtime} represents scenes with explicit anisotropic Gaussian kernels, achieving real-time rendering through a fast alpha-blending splatting algorithm. While 3DGS achieves high rendering quality, it does not strictly constrain surface geometry. To improve surface quality, methods such as SuGaR~\cite{guedon2023sugarsurfacealignedgaussiansplatting}, 2DGS~\cite{Huang2DGS2024}, and GoF~\cite{yu2024gaussianopacityfieldsefficient} combine Gaussians with SDF representations or use more precise intersection calculations to better align with scene surfaces. However, these approaches still require computationally intensive post-processing to extract meshes, which is not integrated into the optimization and may introduce geometric errors.

Methods such as Nvdiffrec~\cite{Munkberg_2022_CVPR} and Flexicubes~\cite{shen2023flexicubes} use implicit SDF representations combined with differentiable iso-surfacing algorithms to enable end-to-end mesh optimization from multi-view images. While these methods produce high-quality meshes that support subsequent geometric processing, their reliance on dense volumetric SDF representations leads to high storage and computational costs. Additionally, they require additional regularization losses to maintain distance-field properties, which can sacrifice fine surface details. In contrast, our method leverages the sparsity of point clouds and the theoretical smoothness guarantees of the IMLS formulation to reconstruct detailed meshes without additional regularization, at a fraction of the computational cost.

\section{Method}\label{sec:method}

\subsection{Implicit Kernel}
%------------------------------------------------%
% Describe input, point clouds, normal
%------------------------------------------------%
Similar to IMLS Splatting~\cite{yang2025imls}, the input to our method is a set of points $\displaystyle \mathcal{P}=\left\{\boldsymbol{p}_i\in \mathbb{R}^3\right\}_{i=1}^N$, obtained from COLMAP~\cite{schoenberger2016sfm}~\cite{pan2024glomap} which lie sufficiently close to the surface $\displaystyle \mathcal{S}$ of a closed, orientable 3D object. For each point $\displaystyle \boldsymbol{p}_i \in \mathcal{P}$, a normal vector $\boldsymbol{n}_i\in \mathbb{R}^3$ is associated, which approximates the surface normal of the underlying surface $\displaystyle \mathcal{S}$. 

%------------------------------------------------%
% Describe MLS algorithm (Kolluri)
%------------------------------------------------%

%------------------------------------------------%
%point function
%------------------------------------------------%
For each sample point  $\displaystyle \boldsymbol{p}_i \in \mathcal{P}$ that lies on the tangent plane of the surface mesh, the point function $\displaystyle\Omega(\boldsymbol{q})$ is a projection operator of $\displaystyle \boldsymbol{q}$ to $\displaystyle \boldsymbol{p}_i$.
\begin{equation}\label{eq:proj}
    \Omega(\boldsymbol{q}) = (\boldsymbol{q}-\boldsymbol{p}_i)\cdot\boldsymbol{n}_i     
\end{equation}
%------------------------------------------------%
 %Describe field
 %------------------------------------------------%
To obtain a three-dimensional SDF function $\displaystyle \mathcal{F}(\boldsymbol{q})$ representing the surface $\displaystyle \mathcal{S}$, the point functions $\displaystyle\Omega(\boldsymbol{q})$ are combined using a kernel, $\displaystyle \gamma(\boldsymbol{q})$.
\begin{equation}\label{eq:sdf}
    \mathcal{F}(\boldsymbol{q}) = \frac{\sum_{\boldsymbol{p}_i \in \mathcal{P}} \gamma(\boldsymbol{q}) \cdot \Omega(\boldsymbol{q})}{\sum_{\boldsymbol{p}_i \in \mathcal{P}} \gamma(\boldsymbol{q})}
\end{equation} 
%------------------------------------------------%
%Decribe how to extract sdf
%------------------------------------------------%
The $0$-level set of $\mathcal{F}$ obtains the reconstructed surface $\displaystyle \mathcal{S}$ as presented in the works by~\citet{kolluri2008mls}.
%------------------------------------------------%
% Describe IMLS splatting
%------------------------------------------------%
In the IMLS Splatting algorithm, the authors used an exponential function for the kernel defined as 
\begin{equation}\label{eq:kernel-imls}
    \gamma_{\textrm{imls}}(\boldsymbol{q}) = \exp{\left\{-\|\boldsymbol{q} - \boldsymbol{p}_i\|^2 / r_i^2\right\}}    
\end{equation}
where $\displaystyle r_i$ is the influence radius.
%------------------------------------------------%
% Describe the role of influence radius
% Refer to kolluri paper for sampling effect and its relation with influence radius
%------------------------------------------------%
It determines the contribution of a specific sample point $p_i$ to the reconstructed surface at any given point $\displaystyle \boldsymbol{q}$. Because the weight decays exponentially, the reconstructed surface $\displaystyle \mathcal{S}$ is primarily determined by samples within a small distance, specifically within a ball of radius $2r_i$ of the evaluation point $\boldsymbol{q}$.

%------------------------------------------------%
% Explain disadvantages of Gaussian Kernel
%------------------------------------------------%
However, the exponential kernel used in~\cite{yang2025imls} inherently has infinite support, meaning its influence extends globally across the spatial domain. In practical computational implementations, this necessitates an arbitrary truncation of the kernel's support radius, such as $\displaystyle r_i$. Such a heuristic cut-off lacks a rigorous mathematical basis. and thus introduces structural artifacts and truncation errors at the boundary limits. Moreover, the exponential function is characterized by a smooth, asymptotic fall-off that remains uniform. This rigid decay profile severely restricts the kernel's ability to adapt dynamically to complex, high-frequency structural variations, thereby limiting the geometric fidelity of the reconstructed surface.

%------------------------------------------------%
%Describe novel kernel
%------------------------------------------------%
To overcome these shortcomings, we propose a novel compact polynomial kernel~\cite{cheng2014kernel} with implicit local support 
\begin{equation}\label{eq:imp_kernel}
    \gamma_{\textrm{ours}}(s_i,k_i,m_i) = 
    \begin{cases} 
        \left(1 - \frac{s_i}{m_i k_i}\right)^{2m_i} \left(\frac{2s_i}{k_i} + 1\right), & \textrm{if} \;\; s_i \in [0,  m_ik_i] \\ 
        0, & \textrm{if} \;\; s_i > m_ik_i 
    \end{cases}
\end{equation}
where $\displaystyle s_i = \|\boldsymbol{q}-\boldsymbol{p}_i\|^2$, and $k_i \,\&\, m_i$ are parameters specific to $\displaystyle i^{\textrm{th}}$ point. As evident from Equation~\eqref{eq:imp_kernel}, the shape of the kernel depends on two parameters, $\displaystyle m_i$ and $\displaystyle k_i$.

\begin{figure*}[t]
    \centering
    \includegraphics[width=\textwidth]{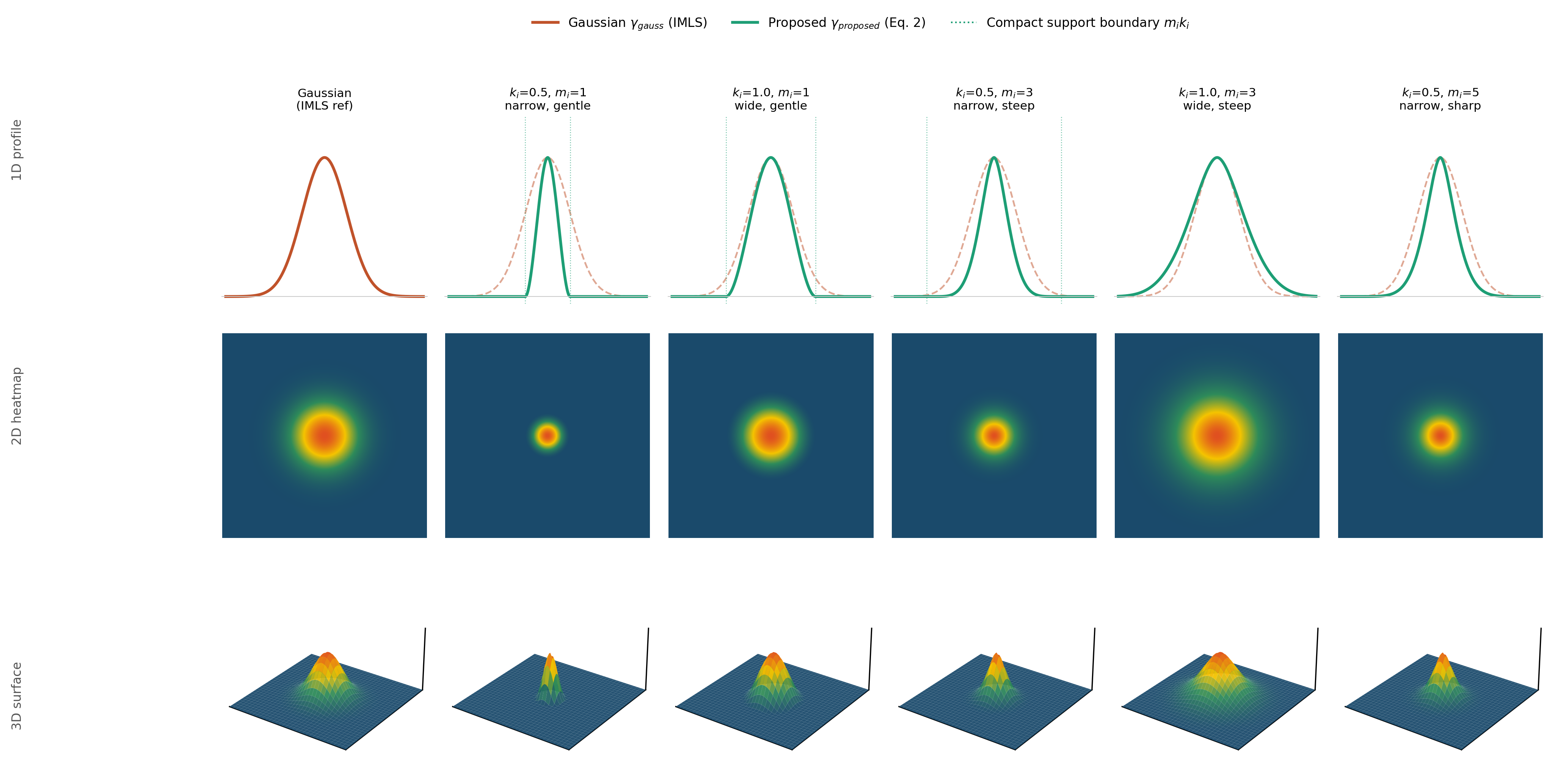}
    \caption{Visualization of the proposed kernel $\displaystyle \gamma_{\text{ours}}$ under varying parameters $\displaystyle k_i$ and $\displaystyle m_i$, compared against the exponential kernel $\displaystyle \gamma_{\textrm{imls}}$ used in IMLS Splatting~\cite{yang2025imls}. The 1-D profiles \emph{(top)} demonstrate that the proposed kernel drops to exactly zero at the compact support boundary $m_i k_i$ (dotted line), unlike the exponential kernel, which extends infinitely and requires arbitrary truncation. The 2-D heatmaps \emph{(middle)} and 3-D surface plots \emph{(bottom)} further illustrate how $\displaystyle k_i$ controls the spatial extent of influence while $\displaystyle m_i$ governs the sharpness of the decay profile, demonstrating the kernel's flexibility in adapting to local surface geometry.}
    \label{fig:kernel_comparison}
\end{figure*}

%------------------------------------------------%
% Describe how the novel kernel is more flexible
% how it is more adaptive
% the role of additional parameters
% more degrees of freedom
%------------------------------------------------%
Our proposed kernel introduces three critical mathematical advantages over the exponential kernel in~\cite{yang2025imls} --- inherent finite support, adaptive parameterization, and additional degrees of freedom. Unlike the standard exponential kernel, which possesses infinite support and must rely on arbitrary, pre-defined distance thresholds to truncate its influence, our formulation naturally decays to zero (see Figure~\ref{fig:kernel_comparison}). This inherent finite support ensures a mathematically rigorous, continuous boundary, effectively eliminating the truncation errors and gradient discontinuities caused by heuristic cutoffs. Furthermore, by introducing additional degrees of freedom through adjustable shape parameters, the proposed formulation enables precise control over the kernel's decay profile, as shown in Figure~\ref{fig:kernel_comparison}. This expanded parameter space enables a much sharper, localized weighting scheme than the rigid asymptotic fall-off of the Gaussian function. Consequently, the kernel can dynamically adapt to local spatial configurations, significantly improving the algorithm's capacity to capture high-frequency surface geometry and preserve intricate, fine-grained structural details that would otherwise be lost to over-smoothing.

%------------------------------------------------%
% explain the feature function
%------------------------------------------------%
We also need to optimize a per-point texture feature $\boldsymbol{c}_i \in \mathbb{R}^D$ for differentiable neural rendering and multi-view optimization. The texture feature is calculated similarly to the signed distance field as
\begin{equation}\label{eq:tex-feature}
    \mathcal{C}(\boldsymbol{q}) = \frac{\sum_{\boldsymbol{p}_i \in \mathcal{P}} \gamma(\boldsymbol{q}) \cdot \boldsymbol{c}_i}{\sum_{\boldsymbol{p}_i \in \mathcal{P}} \gamma(\boldsymbol{v})}
\end{equation}
The texture field is then interpolated onto the extracted surface mesh to model the output texture. Each point is optimized for $\displaystyle 8+D$ parameters $\displaystyle (\boldsymbol{p}_i, \boldsymbol{n}_i, k_i, m_i, \boldsymbol{c}_i)$. The optimization pipeline closely follows the IMLS Splatting~\cite{yang2025imls} algorithm. The influence of each point on the surrounding points differs from one another. We account for this by considering the influence radius as $\displaystyle k_i$ in Equation~\eqref{eq:imp_kernel}. Thus, points with larger $\displaystyle k_i$ are splatted over a broader region. The gradient flow for the backpropagation is derived in Appendix~\ref{app:backprop}. Following the baseline IMLS Splatting framework, we compute the grid-Based splatting, iso-surfacing, SDF generation, and neural shading for the final output. These procedures are explained in Appendix~\ref{app:splat}.

\subsection{Stochastic Preconditioning}
%------------------------------------------------%
% add convolution equation
% describe more about the role of SP
%------------------------------------------------%

% Gaussian points that represent the scene may be trapped in some local minima or collapse into 
% unstable configurations. Also as this heavily resides on gradients from the image reconstruction loss,
%  the gradients could become highly sensitive to exact local positions. 
%  This could lead to slow convergence, unstable training, degenerate point configurations. 
%  To mitigate this, we use stochastic preconditioning where the input gaussian points are perturbed with
%   small random offsets during training. This is just like adding a controlled randomness to the input points. 

%   $p_i$ = $p_i$ + $\epsilon$, where $\epsilon$ is the randomness.

IMLS Splatting optimization can suffer from poor local minima, floater artifacts, and high sensitivity of gradients to exact query positions, particularly in indirectly supervised settings where the only signal comes from image reconstruction losses. To mitigate this, we adopt stochastic preconditioning~\cite{ling2025stoc}, which proposes perturbing the neural query positions in random directions. The effect is similar to smoothening/blurring the neural field by a Gaussian filter.
\begin{equation}
    \textrm{Blur}_\alpha[f](\boldsymbol{q}) = \int_{\Omega} f(\boldsymbol{q} + \boldsymbol{\tau})G(\boldsymbol{\tau})
    = \mathbb{E}\left[f(\boldsymbol{q} + \boldsymbol{\delta})\right],
    \;\; \textrm{s.t.}\;\;\; \boldsymbol{\delta} \sim \mathcal{N}(\mathbf{0}, \alpha \mathbf{I}), \; \alpha \in \mathbb{R}^{+}
    \label{eq:sp_expectation}
\end{equation}
where $\displaystyle G(x) = \left(\sqrt{2\pi}\sigma\right)^{-1}\exp{\left\{-x^2/2\sigma^2\right\}}$ is Gaussian kernel. Querying the updated field during optimization smooths the loss landscape, damps high-frequency gradients, and encourages the field to resolve low-frequency structure before fine detail. Since the grid voxel domain is bounded to $[0,1]^3$ (see Appendix~\ref{app:splat}) and perturbation vectors are unbounded, out-of-bounds coordinates are mapped back via the period-2 reflection operator:
\begin{equation}
    R(x) =
    \begin{cases}
        \textrm{mod}(x, 2), &  \textrm{if } \;\; \textrm{mod}(x, 2) \leq 1, \\
        2 - \textrm{mod}(x, 2), & \textrm{if } \;\; \textrm{mod}(x, 2) > 1,
    \end{cases}
    \label{eq:reflect}
\end{equation}

\subsection{Laplacian Filtering}
%------------------------------------------------%
% add equation 
% make it more descriptive
%------------------------------------------------%
However, stochastic preconditioning smooths the neural field, thereby losing high-frequency features like sharp edges, thin structures, and small surface variations. To recover these, we applied Laplacian filtering, $\displaystyle L = \nabla^2 G $, along with preconditioning done by the Gaussian filter, which measures the local curvature of a function.
\begin{equation}\label{eq:laplacian}
    \begin{aligned}
        \textrm{Lap}_\alpha[f](\boldsymbol{q}) &= \int_{\Omega} f(\boldsymbol{q} + \boldsymbol{\tau})L(\boldsymbol{\tau}) = \int_{\Omega} f(\boldsymbol{q} + \boldsymbol{\tau})\nabla^2 G(\boldsymbol{\tau})\\
        &= \sigma^{-4}\mathbb{E}\left[\delta^2f(\boldsymbol{q}+\boldsymbol{\delta})\right] - \sigma^{-2}\mathbb{E}\left[f(\boldsymbol{q}+\boldsymbol{\delta})\right], \;\; \textrm{s.t.}\;\;\; \boldsymbol{\delta} \sim \mathcal{N}(\mathbf{0}, \alpha \mathbf{I}), \; \alpha \in \mathbb{R}^{+}
    \end{aligned}
\end{equation}

\begin{comment}
\begin{equation}
F(\mathbf{q}) = \frac{A(\mathbf{q})}{B(\mathbf{q})},
\end{equation}
where
\begin{equation}
A(\mathbf{q}) = \sum_i \gamma_i(s_i)\,\Omega_i, 
\quad 
B(\mathbf{q}) = \sum_i \gamma_i(s_i),
\end{equation} 
with
\(
s_i = \|\mathbf{q} - \mathbf{p}_i\|
\)
and
\(
\Omega_i = (\mathbf{q} - \mathbf{p}_i)\cdot \mathbf{n}_i.
\)

\paragraph{Laplacian of the Field.}
Using the quotient rule, the Laplacian of $F$ is given by:
\begin{equation}
\nabla^2 F =
\frac{\nabla^2 A}{B}
- \frac{A\,\nabla^2 B}{B^2}
- \frac{2}{B}(\nabla B \cdot \nabla F).
\end{equation}

Since $\gamma_i$ is a radial function of $s_i$, its derivatives follow:
\begin{equation}
\nabla \gamma_i = \gamma_i'(s_i)\,\hat{\mathbf{r}}_i,
\quad
\nabla^2 \gamma_i = \gamma_i''(s_i) + \frac{2}{s_i}\gamma_i'(s_i),
\end{equation}
where
\(
\hat{\mathbf{r}}_i = \frac{\mathbf{q} - \mathbf{p}_i}{s_i}.
\)

Using these identities, we obtain:
\begin{equation}
\nabla^2 B = \sum_i \left(\gamma_i''(s_i) + \frac{2}{s_i}\gamma_i'(s_i)\right),
\end{equation}
\begin{equation}
\nabla^2 A = \sum_i \left[
\left(\gamma_i''(s_i) + \frac{2}{s_i}\gamma_i'(s_i)\right)\Omega_i
+ 2\,\gamma_i'(s_i)(\hat{\mathbf{r}}_i \cdot \mathbf{n}_i)
\right].
\end{equation}
\end{comment}

After stochastic preconditioning and the Laplacian filter, the neural field becomes
\begin{equation}
    f_{\textrm{final}}(\boldsymbol{q}) = \textrm{Blur}_\alpha[f](\boldsymbol{q}) + \lambda\, \textrm{Lap}_\alpha[f](\boldsymbol{q})
\end{equation}
where $\displaystyle \lambda \in \mathbb{R}^{+}$ controls the strength of the filtering.

% ------------------
% add the derivation
% ------------------

\subsection{Loss Function}
Our architecture is similar to the IMLS splatting~\cite{yang2025imls}, except for the updated kernel, stochastic, and Laplacian regularizers. is trained end-to-end by comparing the rasterized images against the ground-truth multi-view images. Similar to 3DGS~\cite{kerbl20233dgaussiansplattingrealtime} our loss function consists of a pixel-wise $\mathcal{L}_1$ loss and a D-SSIM loss term:
\begin{equation}\label{eq:loss}
    \mathcal{L} = (1 - \lambda)\mathcal{L}_1(I, I_{gt}) + \lambda\mathcal{L}_{ssim}(I, I_{gt})
\end{equation}
where $\lambda$ is empirically set to $0.2$. Background pixels are masked out and randomized during training to reduce ambiguity between foreground and background.

\begin{comment}
To avoid local minima and progressively recover high-frequency details, we employ the dynamic point resampling and upsampling strategy described in the IMLS Splatting. 
\begin{itemize}
    \item \textbf{Resampling:} We periodically evaluate the visibility of the reconstructed mesh triangles. Triangles unobserved by any training camera are discarded. We then sample a new set of points at the centroids of the remaining valid triangles, assigning them the averaged texture features and normals of their respective triangles. This ensures the point cloud remains evenly distributed and strictly adheres to the optimized surface.
    \item \textbf{Grid Upsampling:} To facilitate a coarse-to-fine reconstruction, we progressively increase the spatial resolution of the 3D splatting grid. A finer grid yields a denser mesh via Marching Cubes, which in turn generates a higher-density point cloud during the resampling phase, allowing the field to capture increasingly finer local details.
\end{itemize}
\end{comment}

\section{Experiments}\label{sec:experiments}
In this section, we evaluate the performance of our proposed method
focusing on two primary tasks --- high-fidelity surface reconstruction and novel view synthesis. We provide a comprehensive breakdown of our implementation environment, followed by a comparative analysis, both quantitative and qualitative, against current state-of-the-art (SOTA) baselines, including the original implementation of IMLS Splatting~\cite{yang2025imls}. Finally, we conduct ablation studies to isolate and validate the specific contributions of our novel kernel design, as presented in Appendix~\ref{app:ablation}.

\subsection{Implementation}
All experiments are conducted on a single NVIDIA RTX 6000 GPU using PyTorch. Two standard datasets, NeRF-synthetic~\cite{mildenhall2020nerfrepresentingscenesneural} and DTU~\cite{jensen2014large}, are used for evaluation. When evaluating the model's performance on both the NeRF Synthetic and DTU datasets, a grid resolution of $256^3$ yields a detailed reconstruction, with an average of 471k points and a face count of 849k per iteration. This process requires a GPU compute memory allocation of $11.0$ GB. 

%Despite the geometric density, the system maintains impressive temporal efficiency: the IMLS splatting phase (with our novel kernel) averages 4.82 ms per iteration, the Marching cubes extraction completes in 3.89 ms per iteration, while rasterization takes 3.01 ms and shading takes 6.94 ms per iteration. 

\subsection{Surface Reconstruction}
We evaluate the surface reconstruction quality of our method by benchmarking it against a comprehensive suite of state-of-the-art approaches. These include implicit representations such as VolSDF~\cite{yariv2021volume}, NeuS~\cite{wang2021neus}, and Neuralangelo~\cite{li2023neuralangelohighfidelityneuralsurface}, Gaussian-based frameworks like SuGaR~\cite{guedon2023sugarsurfacealignedgaussiansplatting}, 2DGS~\cite{Huang2DGS2024}, and GoF~\cite{yu2024gaussianopacityfieldsefficient}, and explicit mesh-based methods such as Nvdiffrec~\cite{Munkberg_2022_CVPR}, Flexicubes~\cite{shen2023flexicubes} and IMLS-Splatting~\cite{yang2025imls}.

To ensure a rigorous and fair evaluation, we adopt the standardized settings utilized in recent literature. For mesh-based approaches that typically require object masks (e.g., Nvdiffrec and Flexicubes), we compare against their manifold settings without vertex refinement. For implicit and Gaussian-based methods, we include results both with and without mask-based supervision ($\dagger$) to highlight our method’s robustness. While IMLS Splatting originally improved upon these baselines by randomly assigning background colors to avoid mask dependency, our work focuses on enhancing the underlying reconstruction kernel to further reduce loss in high-fidelity details.

Table~\ref{tab:dtu_comparison} summarizes the quantitative results on the DTU real-world dataset using the Chamfer Distance (CD) metric. Our method, powered by the novel kernel and filtering techniques, achieves an average CD of $0.41$, outperforming the original IMLS Splatting ($0.57$) and establishing a new state-of-the-art for this framework. The efficiency of our approach is equally significant, as it bridges the gap between the rapid training of Gaussian-based methods and the geometric precision of implicit surface representations. While implicit methods such as VolSDF and NeuS often require over 12 hours of training to converge, our kernel-driven splatting approach yields superior or comparable geometric results in only 11 minutes. Although 2DGS offers a slightly faster training time of 9 minutes, our method provides a much higher level of surface detail and lower error rates, making it a more robust choice for high-quality surface extraction tasks~\cite{li2023neuralangelohighfidelityneuralsurface}. The reduction in CD across most scenes demonstrates that our kernel better captures fine-grained surface details and reduces noise in the reconstructed mesh compared to standard IMLS formulations.

% Define the colors used
\definecolor{best}{RGB}{255, 200, 200}    % Reddish-pink
\definecolor{second}{RGB}{255, 230, 200}  % Orange/Peach
\definecolor{third}{RGB}{255, 255, 200}   % Light Yellow

\begin{table*}[t]
\centering
\caption{Comparison of Chamfer Distance on DTU dataset for mesh quality evaluation. We compare with implicit methods, Gaussian-based methods, and mesh-based methods. $\dagger$ indicates that object masks were utilized during training, and $*$ indicates the reproduced result using the official code. All 3DGS-based methods and ours start from COLMAP points, as we use early-stage coarse 3DGS points as initialization. We color each cell as best \colorbox{best}{ }, second best \colorbox{second}{ }, and third best \colorbox{third}{ }.}

\label{tab:dtu_comparison}

\resizebox{\textwidth}{!}{%
\begin{tabular}{l|ccccccccccccccc|c|c}
\toprule
 & \textbf{24} & \textbf{37} & \textbf{40} & \textbf{55} & \textbf{63} & \textbf{65} & \textbf{69} & \textbf{83} & \textbf{97} & \textbf{105} & \textbf{106} & \textbf{110} & \textbf{114} & \textbf{118} & \textbf{122} & \textbf{Avg.} & \textbf{Time} \\ \midrule
 
VolSDF~\cite{yariv2021volume} & 1.14 & 1.26 & 0.81 & 0.49 & 1.25 & 0.70 & 0.72 & 1.29 & 1.18 & 0.70 & 0.66 & 1.08 & 0.42 & 0.61 & 0.55 & 0.86 & >12h \\

NeuS~\cite{wang2021neus} & 1.00 & 1.37 & 0.93 & 0.43 & 1.10 & 0.65 & 0.57 & 1.48 & 1.09 & 0.83 & 0.52 & 1.20 & 0.35 & 0.49 & 0.54 & 0.84 & >12h \\

Neuralangelo~\cite{li2023neuralangelohighfidelityneuralsurface} & \cellcolor{best}0.37 & \cellcolor{second}0.72 & 0.35 & 0.35 & \cellcolor{third}0.87 & \cellcolor{third}0.54 & \cellcolor{third}0.53 & 1.29 & 0.97 & 0.73 & 0.47 & 0.74 & \cellcolor{second}0.32 & 0.41 & 0.43 & \cellcolor{third}0.61 & >12h \\
\midrule
SuGaR~\cite{guedon2023sugarsurfacealignedgaussiansplatting} & 1.47 & 1.33 & 1.13 & 0.61 & 2.25 & 1.71 & 1.15 & 1.63 & 1.62 & 1.07 & 0.79 & 2.45 & 0.98 & 0.88 & 0.79 & 1.33 & 52m \\
2DGS~\cite{Huang2DGS2024} & 0.48 & 0.91 & 0.39 & 0.39 & 1.01 & 0.83 & 0.81 & 1.36 & 1.27 & 0.76 & 0.70 & 1.40 & 0.40 & 0.76 & 0.52 & 0.80 & \cellcolor{best}9m \\

GoF~\cite{yu2024gaussianopacityfieldsefficient} & 0.50 & 0.82 & 0.37 & 0.37 & 1.12 & 0.74 & 0.73 & 1.18 & 1.29 & 0.68 & 0.77 & 0.90 & 0.42 & 0.66 & 0.49 & 0.74 & \cellcolor{third}18m \\ \midrule

Neus$\dagger$~\cite{wang2021neus} & 0.83 & 0.98 & 0.56 & 0.37 & 1.13 & 0.59 & 0.60 & 1.45 & \cellcolor{third}0.95 & 0.78 & 0.52 & 1.43 & 0.36 & 0.45 & 0.45 & 0.77 & >12h \\

Neuralangelo$\dagger$~\cite{li2023neuralangelohighfidelityneuralsurface} & \cellcolor{second}0.45 & \cellcolor{third}0.74 & \cellcolor{third}0.33 & \cellcolor{second}0.34 & 1.05 & \cellcolor{third}0.54 & \cellcolor{third}0.53 & 1.33 & 1.05 & 0.72 & \cellcolor{third}0.43 & \cellcolor{third}0.69 & 0.34 & \cellcolor{third}0.38 & \cellcolor{third}0.42 & 0.62 & >12h \\

2DGS$\dagger$~\cite{Huang2DGS2024} & \cellcolor{third}0.46 & 0.84 & \cellcolor{second}0.31 & 0.45 & 0.92 & 1.01 & 0.83 & \cellcolor{second}1.23 & 1.30 & \cellcolor{third}0.66 & 0.61 & 1.07 & 0.45 & 0.71 & 0.54 & 0.76 & \cellcolor{best}9m \\ 

Nvdiffrec$\dagger$~\cite{Munkberg_2022_CVPR} & 3.04 & 3.02 & 2.10 & 0.78 & 2.18 & 1.60 & 1.46 & 1.67 & 2.85 & 1.26 & 1.10 & 3.26 & 1.13 & 1.31 & 1.19 & 1.86 & >1h \\

Flexicubes $\dagger$~\cite{shen2023flexicubes} & 1.66 & 1.60 & 0.91 & 0.50 & 2.60 & 1.32 & 0.87 & 1.45 & 1.60 & 1.38 & 0.73 & 1.85 & 1.01 & 0.64 & 0.75 & 1.26 & >1h \\

IMLS Splatting$\dagger\ast$~\cite{yang2025imls}* & 0.52 & 1.02 & 0.37 & \cellcolor{third}0.32 & \cellcolor{second}0.86 & \cellcolor{second}0.50 & \cellcolor{second}0.48 & \cellcolor{third}1.15 & \cellcolor{second}0.76 & \cellcolor{second}0.59 & \cellcolor{second}0.37 & \cellcolor{second}0.67 & \cellcolor{second}0.33 & \cellcolor{third}0.33 & \cellcolor{second}0.34 & \cellcolor{second}0.57 & \cellcolor{second}11m \\

\textbf{Ours$\dagger\ast$} & 0.50 & \cellcolor{best}0.66 & \cellcolor{best}0.27 & \cellcolor{best}0.20 & \cellcolor{best}0.57 & \cellcolor{best}0.34 & 
\cellcolor{best} 0.34 &
\cellcolor{best}0.81 & \cellcolor{best}0.70 & \cellcolor{best}0.35 & \cellcolor{best}0.28 & \cellcolor{best}0.41 &  \cellcolor{best}0.24 & \cellcolor{best}0.22 & \cellcolor{best}0.21 & \cellcolor{best}0.41 & \cellcolor{second}11m \\ \bottomrule
\end{tabular}%
}
\end{table*}
% ---------------------------------------------------- %
% Visual comparison figure
% ---------------------------------------------------- %
\begin{figure}[h]
    \centering
    \includegraphics[width=\textwidth]{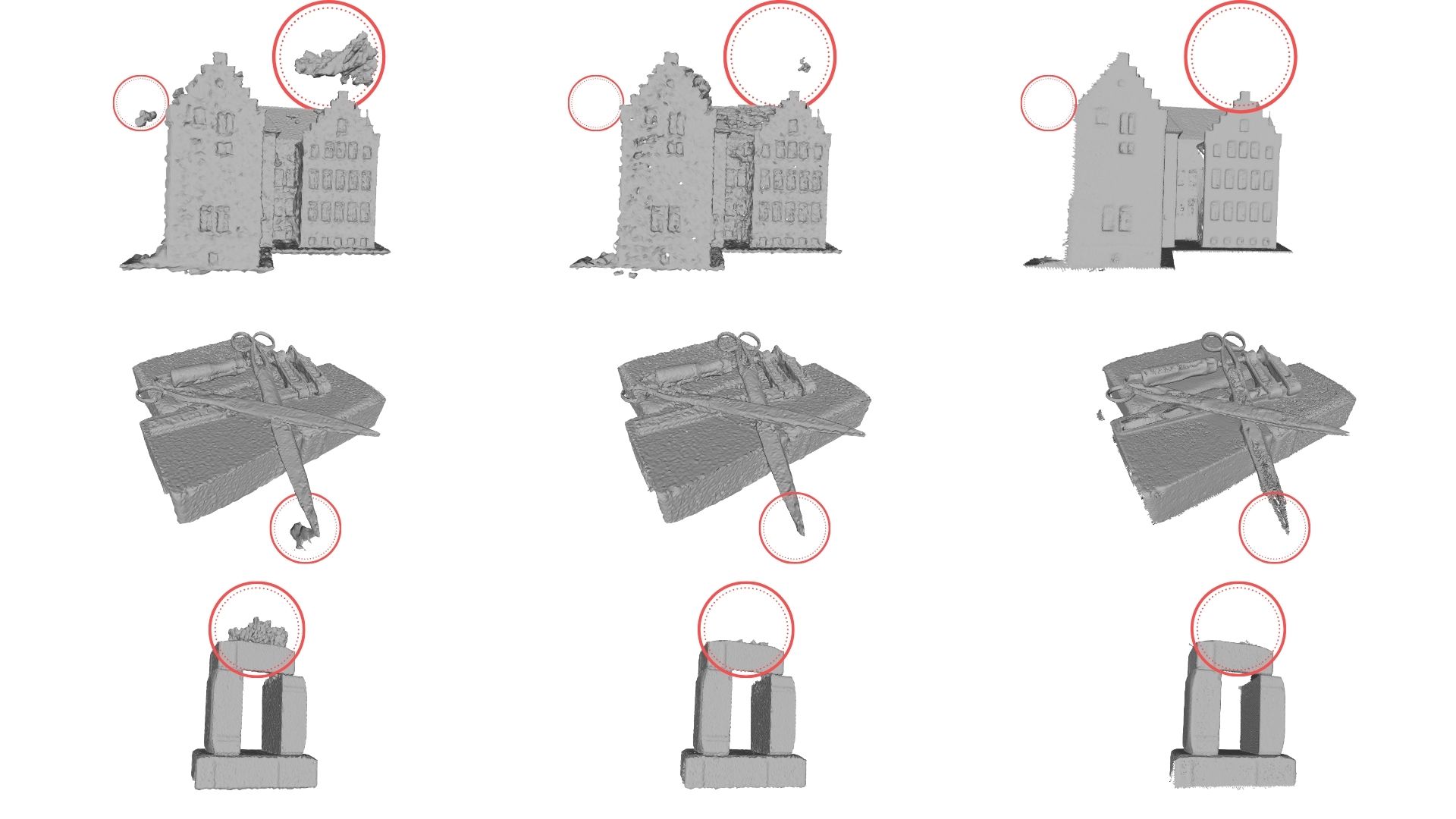} 
    \caption{A qualitative comparison of the generated meshes. Meshes generated using IMLS Splatting~\cite{yang2025imls} are on the left, meshes generated using our method are in the middle and the reference meshes are on the right. In contrast to IMLS Splatting, our approach consistently reduces or eliminates geometric artifacts, yielding meshes that align more closely with the reference geometry.}
    \label{fig:mesh_comparison}
\end{figure}
Fig.~\ref{fig:mesh_comparison} presents a visual comparison of surface reconstruction on DTU dataset. Our proposed method significantly improves the visual quality of the reconstructed meshes, effectively suppressing geometric artifacts prevalent in standard splatting-based approaches, particularly in challenging regions with sparse view coverage. While IMLS Splatting often introduces noticeable geometric noise, such as the spurious floating structures and unconstrained boundary extensions highlighted by red circles in Figure ~\ref{fig:mesh_comparison}, our novel kernel formulation provides a superior regularization effect, effectively suppressing these spurious structures. By more accurately modeling the spatial influence of the splats, our approach maintains a clean, continuous surface even in low-data areas. This results in reconstructions that are not only visually smoother but also more faithful to the ground-truth reference geometry, as seen in the second and third columns of Figure ~\ref{fig:mesh_comparison}.

\subsection{Novel-View Synthesis}
\begin{figure}[h!tb]
    \centering
    \includegraphics[width=\textwidth]{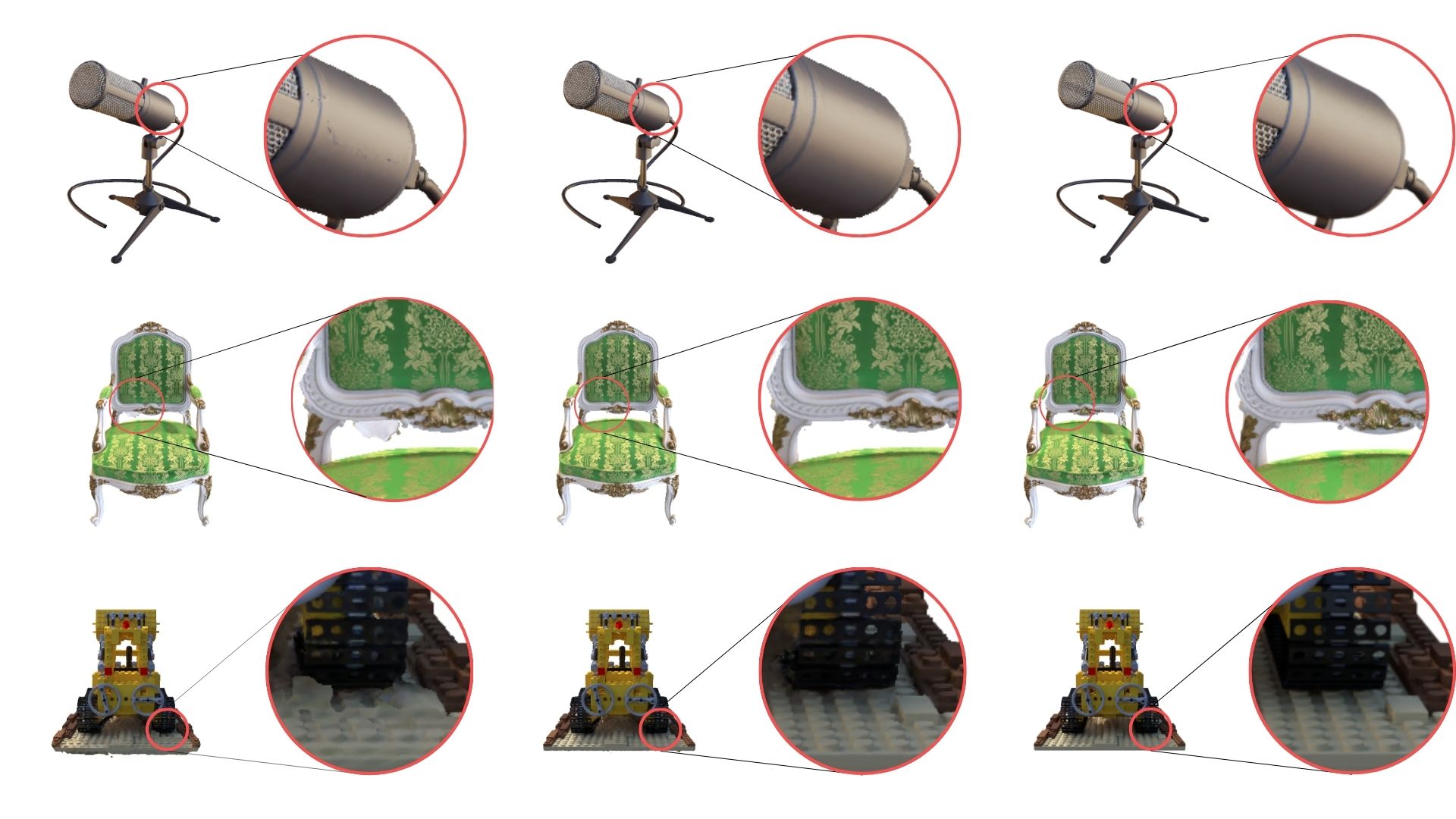} 
    \caption{Qualitative rendering comparison with IMLS Splatting~\cite{yang2025imls} method. Images generated using the IMLS Splatting method are on the left, images generated using our method are in the middle, and the reference meshes are on the right. Our method demonstrates noticeably sharper details and fewer geometric artifacts in complex scenes.}
    \label{fig:rendering_comparison_image}
\end{figure}

We utilize Supersample Anti-Aliasing (SSAA) during evaluation to mitigate rasterization artifacts.
Fig.~\ref{fig:rendering_comparison_image} demonstrates that our kernel yields sharper textures on complex patterns with the NeRF Synthetic dataset~\cite{mildenhall2020nerfrepresentingscenesneural}. 
% As shown in Table~\ref{tab:rendering_quality_final} and Table~\ref{tab:rendering_comparison_detailed}, our method outperforms IMLS Splatting in PSNR, SSIM, LPIPS, and MSE on the NeRF-synthetic dataset while preserving high-frequency details.  
Our improved formulation achieves competitive performance compared to IMLS Splatting, consistently reaching parity or showing slight improvements across the majority of the NeRF-synthetic dataset, as shown in Table~\ref{tab:rendering_quality_final} and Table~\ref{tab:rendering_comparison_detailed}. While the quantitative deltas are subtle, our method outperforms the baseline in the average values of PSNR, LPIPS, and MSE. Please note that all these values are obtained by running the open-source codebase released by the authors of IMLS Splattings~\cite{yang2025imls}.

% ------------------------------------------------------ %
% Quantitative results
% ------------------------------------------------------ %
\definecolor{best}{RGB}{255, 178, 178}
\begin{table*}[htbp]
\centering
\scriptsize
\caption{Quantitative rendering comparison on the NeRF-synthetic dataset~\cite{mildenhall2020nerfrepresentingscenesneural}. We compare the IMLS Splatting~\cite{yang2025imls} against our method across eight scenes. Please note that we run the open-source code released by IMLS Splatting to obtain the values. Cells highlighted in \colorbox{best}{} indicate the superior performance for each metric per scene.}
\label{tab:rendering_quality_final}
\resizebox{\textwidth}{!}{%
\begin{tabular}{ll|cccccccc|c}
\toprule
\textbf{Metric} & \textbf{Method} & \textbf{Lego} & \textbf{Chair} & \textbf{Drums} & \textbf{Ficus} & \textbf{Hotdog} & \textbf{Mic} & \textbf{Materials} & \textbf{Ship} & \textbf{Avg.} \\ \midrule

\multirow{2}{*}{PSNR $\uparrow$} & IMLS & 28.9987
 & 32.2378 & \cellcolor{best}23.27662 &\cellcolor{best} 26.0216 & 29.0326 & 30.2919 & 27.0560 & 16.8776 & 26.7240 \\
 
 & \textbf{Ours} & \cellcolor{best}29.7835 & \cellcolor{best}32.8831 & 22.8156 & 25.9323 & \cellcolor{best}29.8151 & \cellcolor{best}30.4122 & \cellcolor{best}27.1354 & \cellcolor{best}16.105 & \cellcolor{best}26.8602 \\ \midrule

\multirow{2}{*}{SSIM $\uparrow$} & IMLS & 0.9619 & 0.9767 & \cellcolor{best}0.9260 & \cellcolor{best}0.9508 & 0.9730 & 0.9815 & 0.9433 & \cellcolor{best}0.8137 & \cellcolor{best}0.9408 \\

 & \textbf{Ours} & \cellcolor{best}0.9649 & \cellcolor{best}0.9787 & 0.9245 & 0.9503 & \cellcolor{best}0.9740 & \cellcolor{best}0.9820 & \cellcolor{best}0.9439 & 0.8057 & 0.9405 \\ \midrule

\multirow{2}{*}{LPIPS $\downarrow$} & IMLS & 0.0305 & 0.0177 & \cellcolor{best}0.0622 & \cellcolor{best}0.0540 & 0.0589 & \cellcolor{best}0.0331 & 0.0461 & \cellcolor{best}0.1700 & 0.0590 \\

 & \textbf{Ours} & \cellcolor{best}0.0287 & \cellcolor{best}0.0164 & 0.0748 & 0.0585 & \cellcolor{best}0.0321 & 0.0335 & \cellcolor{best}0.0450 & 0.1820 & \cellcolor{best}0.0588 \\ \midrule

\multirow{2}{*}{MSE $\downarrow$} & IMLS & 0.00155 & 0.00063 & \cellcolor{best}0.00511 & 0.00260 & 0.00179 & 0.00095 & 0.00214 & 0.02234 & 0.00463 \\
 & \textbf{Ours} & \cellcolor{best}0.00116 & \cellcolor{best}0.00054 & 0.00534 & \cellcolor{best}0.00259 & \cellcolor{best}0.00128 & \cellcolor{best}0.00092 & \cellcolor{best}0.00120 & \cellcolor{best}0.02021 & \cellcolor{best}0.00415 \\ 
\bottomrule
\end{tabular}%
}
\end{table*}

\begin{comment}
\begin{table}[htbp]
\centering
\caption{Rendering quality comparison on the NeRF-synthetic dataset~\cite{mildenhall2020nerfrepresentingscenesneural}. We compare our approach against standard Gaussian splatting (3DGS, 2DGS) and established mesh-based baselines.}
\label{tab:rendering_comparison_detailed}
\small % Keeps text legible without stretching it
\begin{tabular}{l|ccc}
\toprule
\textbf{Method} & \textbf{PSNR} $\uparrow$ & \textbf{SSIM} $\uparrow$ & \textbf{LPIPS} $\downarrow$ \\ \midrule
3DGS-7k~\cite{kerbl20233dgaussiansplattingrealtime} & 31.29 & 0.96 & 0.04 \\
3DGS~\cite{kerbl20233dgaussiansplattingrealtime} & 33.33 & 0.97 & 0.03 \\
2DGS~\cite{Huang2DGS2024} & 33.06 & 0.97 & 0.03 \\ \midrule
Nvdiffrec~\cite{Munkberg_2022_CVPR} & 26.87 & 0.93 & 0.09 \\
Flexicubes~\cite{shen2023flexicubes} & 27.50 & 0.93 & 0.08 \\ \midrule
IMLS~\cite{yang2025imls} & 26.72 & 0.94 & 0.06 \\ 
\textbf{Ours} & 26.86 & 0.94 & 0.06 \\ \bottomrule
\end{tabular}
\end{table}
\end{comment}

\begin{table}[t]
\centering
\caption{Rendering quality comparison on the NeRF-synthetic dataset~\cite{mildenhall2020nerfrepresentingscenesneural}. We compare our approach against standard Gaussian splatting (3DGS, 2DGS) and established mesh-based baselines.}
\label{tab:rendering_comparison_detailed}

\normalsize
\setlength{\tabcolsep}{2.5pt}
\renewcommand{\arraystretch}{0.95}

\resizebox{0.48\textwidth}{!}{%
\begin{tabular}{lccc|ccc|c}
\toprule
\textbf{Metric} 
& \textbf{3DGS-7k}
& \textbf{3DGS}
& \textbf{2DGS}
& \textbf{Nvdiffrec}
& \textbf{Flexicubes}
& \textbf{IMLS}
& \textbf{Ours} \\
\midrule
PSNR $\uparrow$
& 31.29 & 33.33 & 33.06
& 26.87 & 27.50 & 26.72
& \textbf{26.86} \\

SSIM $\uparrow$
& 0.96 & 0.97 & 0.97
& 0.93 & 0.93 & 0.94
& \textbf{0.94} \\

LPIPS $\downarrow$
& 0.04 & 0.03 & 0.03
& 0.09 & 0.08 & 0.06
& \textbf{0.06} \\
\bottomrule
\end{tabular}
}
\end{table}

We further evaluate our rendering performance against Gaussian-based splatting and state-of-the-art view synthesis methods, as summarized in Table~\ref{tab:rendering_comparison_detailed}. For a comprehensive reference, we include results for 3DGS-7k (the coarse initialization of 3DGS), the full 3DGS~\cite{kerbl20233dgaussiansplattingrealtime}, and 2DGS~\cite{Huang2DGS2024}. Gaussian-based methods like 3DGS generally yield higher PSNR due to their unconstrained volumetric representation. Our method prioritizes geometric integrity while significantly narrowing the appearance gap.

Our method produces visually superior results with fewer artifacts and sharper details, which standard metrics do not fully capture. Despite the numerical similarity in Table \ref{tab:rendering_comparison_detailed}, our method demonstrates a clear reduction in artifacts. As evidenced in Figure~\ref{fig:rendering_comparison_image}, our approach significantly reduces geometric floaters in complex structures, such as the ornate trim of the Chair, and recovers sharp, high-frequency textures in the Lego scene --- like individual baseplate studs --- which are blurred in the IMLS Splatting baseline.

Finally, it is evident from the qualitative and quantitative experimental results that our method excels at maintaining structural integrity and preserving high-frequency details in mesh reconstruction, especially in regions where existing reconstruction methods typically fail. 
\section{Conclusion and Limitations}\label{sec:conclusion}
We presented an enhanced implicit moving least squares surface reconstruction method built upon the IMLS Splatting framework, introducing three focused contributions. We proposed a novel implicit kernel with inherent finite support and adaptive per-point parameters, replacing the original IMLS kernel to eliminate truncation artifacts and enable more precise local surface adaptation. We further integrated stochastic preconditioning to stabilize the optimization trajectory by smoothing the loss landscape during early training, and incorporated Laplacian filtering as a residual correction to the implicit field to recover fine-grained geometric details that are otherwise suppressed by the stochastic preconditioning.

All three modifications integrate seamlessly into the existing grid-based splatting pipeline without altering its core architecture, preserving the computational efficiency and differentiable rendering capability of the original framework. The proposed kernel's mathematically rigorous finite support, combined with its additional degrees of freedom, enables the reconstructed surface to better capture high-frequency geometry, while stochastic preconditioning and Laplacian filtering collectively address the stability and detail-recovery challenges inherent to IMLS-based optimization.

Even though our method generates the surface mesh more accurately, it suffers in visual accuracy, measured by PSNR, SSIM, and LPIPS, compared to NeRF~\cite{mildenhall2020nerfrepresentingscenesneural} or 3DGS~\cite{kerbl20233dgaussiansplattingrealtime}. Moreover, our method does not include translucent rendering. These could be interesting avenues for future research.

% \section*{References}
\bibliographystyle{plainnat}
\bibliography{10_bibliography}

\appendix
\section{Gradient Flow Computation for Backpropagation}\label{app:backprop}
\subsection{Differentiation with respect to $s$} 
\begin{align}
\frac{d\gamma(s)}{ds} &= \frac{d}{ds} \left[ \left(1 - \frac{s}{mk}\right)^{2m} \left(\frac{2s}{k} + 1\right) \right] \nonumber \\ 
&= \left[ -\frac{2}{k} \left(1 - \frac{s}{mk}\right)^{2m-1} \left(\frac{2s}{k} + 1\right) \right] + \left[ \left(1 - \frac{s}{mk}\right)^{2m} \left(\frac{2}{k}\right) \right] \nonumber \\
\frac{d\gamma(s)}{ds} &= \frac{2}{k} \left[ \left(1 - \frac{s}{mk}\right)^{2m} - \left(1 - \frac{s}{mk}\right)^{2m-1} \left(\frac{2s}{k} + 1\right) \right]
\end{align}

\subsection{Differentiation with respect to $k$}
Substituting $\displaystyle \frac{1}{k} = t$, we get $\displaystyle \gamma(s) = \left(1 - \frac{st}{m}\right)^{2m} (2st + 1)$. Using the chain rule
\begin{equation}
    \frac{d\gamma(s)}{dk} = \frac{d\gamma(s)}{dt} \times \frac{dt}{dk}
\end{equation}

\begin{equation}
\begin{aligned}
    &\frac{d\gamma(s)}{dt} = -2s \left(1 - \frac{st}{m}\right)^{2m-1} (2st + 1) + 2s \left(1 - \frac{st}{m}\right)^{2m} \\
    &\textrm{and}\\
    &\frac{dt}{dk} = \frac{d(k^{-1})}{dk} = -\frac{1}{k^2}\\ 
    &\frac{d\gamma(s)}{dk} = -\frac{1}{k^2} \left[ -2s \left(1 - \frac{st}{m}\right)^{2m-1} (2st + 1) + 2s \left(1 - \frac{st}{m}\right)^{2m} \right]\\
    &\textrm{Substituting } t = \frac{1}{k} \textrm{ back in and factoring out } 2s,\\
    &\frac{d\gamma(s)}{dk} = \frac{2s}{k^2} \left[ \left(1 - \frac{s}{mk}\right)^{2m-1} \left(\frac{2s}{k} + 1\right) - \left(1 - \frac{s}{mk}\right)^{2m} \right]
\end{aligned}
\end{equation} 

\subsection{Differentiation with respect to $m$}
Let $\displaystyle y = \left(1 - \frac{s}{mk}\right)^{2m} \left(\frac{2s}{k} + 1\right)$. Applying $\displaystyle \log$ to both sides,
\begin{equation}
    \ln y = \ln\left(\frac{2s}{k} + 1\right) + 2m \ln\left(1 - \frac{s}{mk}\right)
\end{equation}
Differentiating with respect to $m$:
\begin{equation}
\begin{aligned}
    &\frac{1}{y} \frac{dy}{dm} = 2 \ln\left(1 - \frac{s}{mk}\right) + 2m \left( \frac{1}{1 - \frac{s}{mk}} \right) \left( \frac{s}{km^2} \right) \\
    &\frac{dy}{dm} = y \left[ 2 \ln\left(1 - \frac{s}{mk}\right) + \frac{2s}{km \left(1 - \frac{s}{mk}\right)} \right]\\ 
    &\textrm{Substituting } y \textrm{ back to the expression the final derivative of } \gamma(s) \textrm{ with respect to } m, \textrm{ we get }\\
    &\frac{d\gamma(s)}{dm} = \left[ \left(1 - \frac{s}{mk}\right)^{2m} \left(\frac{2s}{k} + 1\right) \right] \times 2 \left[ \ln\left(1 - \frac{s}{mk}\right) + \frac{s}{km \left(1 - \frac{s}{mk}\right)} \right]
\end{aligned}
\end{equation}

These gradients are then passed on to the attributes of the input points.

\section{Grid-based Splatting, SDF Generation and Shading}\label{app:splat}
\subsection{Fast Grid-Based Splatting}
Following the baseline IMLS Splatting framework~\cite{yang2025imls}, we compute the implicit field values on a discrete 3-D grid rather than relying on a computationally expensive query-based approach. Query-based methods compute the field value by searching for neighboring points within a fixed radius, which can introduce irrelevant points or overlook key geometry, leading to discontinuities. Instead, the attributes of each input point are directly splatted onto the grid vertices that fall within its local neighborhood. In our formulation, this neighborhood is explicitly bounded by the finite support of our proposed kernel parameter $\displaystyle k_i$. To maximize computational efficiency, a parallelized GPU Radix sort~\cite{satish2009gpuradix} is employed to associate grid vertices with their influencing points. This enables the parallel computation of both the SDF and the neural texture field across the grid. During the backward pass, the gradients accumulated at the grid vertices are efficiently propagated back to the $\displaystyle 8+D$ trainable parameters of the input point cloud.

\subsection{Iso-Surfacing and Background SDF Regularization}
After SDF and texture features are computed on the 3-D grid, we use the Marching Cubes algorithm~\cite{wei2025neumanifold} to extract the explicit surface mesh and interpolate the corresponding texture features. During the early stages of optimization, sparse point coverage and noisy estimated normals can result in topological artifacts, such as truncated open surfaces or surfaces perpendicular to the actual geometry. To ensure the generation of a stable, watertight mesh, we dilate a background positive distance field around the sparsely covered grid~\cite{yang2025imls}. Because our pipeline extracts the zero-level set, this background SDF does not interfere with valid surface regions but effectively closes arbitrary boundaries, converting problematic open surfaces into outward-facing, watertight geometry that is much easier to optimize.

\subsection{Differentiable Neural Shading}
To drive the multi-view optimization process, we rasterize the extracted mesh into the target camera views using a differentiable rasterizer~\cite{Laine2020diffrast}. This produces a foreground mask, a normal map, and a 2-D feature map for each view. Notably, we compute vertex normals by averaging the normals of surrounding triangles and rasterize them directly, explicitly coupling the mesh geometry to the rendering loss.

To model view-dependent appearance, the rasterized feature maps are processed through a neural shading pipeline. A Multi-Layer Perceptron (MLP) decodes the features into spatially varying scene properties, including diffuse color, specular tint, and specular features. A secondary, lightweight MLP is then used to predict the final specular color based on the reflection direction and the camera viewing direction. This disentanglement allows the model to efficiently capture complex, view-dependent effects such as inter-reflections and occlusions. The process is similar to the one presented in IMLS Splatting~\cite{yang2025imls}.
\section{Ablation Study}\label{app:ablation}
We analyze the contribution of stochastic preconditioning (SP) and Laplacian filtering (Lap) to the reconstruction quality. Specifically, we compare the method without these components to the full model that incorporates both.

\begin{table*}[htbp]
\centering
\caption{Ablation study on the NeRF-synthetic dataset. We evaluate the incremental contribution 
of Stochastic Preconditioning (SP) and Laplacian Filtering (LF) on top of our Improved IMLS 
(proposed kernel). Cells highlighted in \colorbox{best}{} indicate the superior performance 
for each metric per scene.}
\label{tab:ablation}
\resizebox{\textwidth}{!}{%
\begin{tabular}{ll|cccccccc|c}
\toprule
\textbf{Metric} & \textbf{Method} & \textbf{Lego} & \textbf{Chair} & \textbf{Drums} & \textbf{Ficus} & \textbf{Hotdog} & \textbf{Mic} & \textbf{Materials} & \textbf{Ship} & \textbf{Avg.} \\ \midrule

\multirow{3}{*}{PSNR $\uparrow$}
 & Improved IMLS (Proposed Kernel) & 29.5326 & \cellcolor{best}32.8831 & 22.7060 & \cellcolor{best}25.9323 & \cellcolor{best}29.8151 & 30.2500 & 27.0918 & 16.0805 & 26.7864 \\
 & \quad + Stochastic Preconditioning (SP)  & 29.7495 & 32.8036 & 21.9662 & 25.8124 & 29.6015 & 30.3093 & 27.1197 & 16.0759 & \\
  & \quad + Laplacian Filtering (LF)  & 29.6608 & 32.8627 & 22.7986 & 25.8901 & 29.6450 & 30.3105 & 27.0613 & 15.9937 & \\
 & \quad + Stochastic Preconditioning (SP) + Laplacian Filtering (LF) & \cellcolor{best}29.7835 &  32.7298 &  \cellcolor{best}22.8156 & 25.8213  & 29.8130
  &   \cellcolor{best} 30.4112 &  \cellcolor{best}27.1354 &    \cellcolor{best}16.105 & \cellcolor{best}26.8269
  \\ \midrule

\multirow{3}{*}{SSIM $\uparrow$}
 & Improved IMLS (Proposed Kernel) & 0.9628 & \cellcolor{best}0.9787 & \cellcolor{best}0.9245 & \cellcolor{best}0.9503 & \cellcolor{best}0.9740 & 0.9815 & 0.9434 & \cellcolor{best}0.8057 & \cellcolor{best}0.9401 \\
 & \quad + Stochastic Preconditioning (SP)  & 0.9648 & 0.9784 & 0.9215 & 0.9491 & 0.9733 & 0.9817 & 0.9432 & 0.8041 & \\
  & \quad + Laplacian Filtering (LF)  & 0.9645 & 0.9787 & 0.9239 & 0.9496 & 0.9734 & 0.9818 & 0.9431 & 0.8061 & \\
 & \quad + Stochastic Preconditioning (SP) + Laplacian Filtering (LF) & \cellcolor{best}0.9649 & 0.9785 & 0.9242 &  0.9490 & 0.9737 & \cellcolor{best}0.9820 & \cellcolor{best}0.9439 & 0.8043 & 0.9401\\ \midrule

\multirow{3}{*}{LPIPS $\downarrow$}
 & Improved IMLS (Proposed Kernel) & 0.0303 & \cellcolor{best}0.0164 & 0.0756 & \cellcolor{best}0.0585 & \cellcolor{best}0.0321 & 0.0347 & 0.0460 & 0.1823 & 0.0595 \\
 & \quad + Stochastic Preconditioning (SP)  & 0.0293 & 0.0167 & 0.0783 & 0.0603 & 0.0332 & 0.0353 & 0.0453 & 0.1819 & \\
  & \quad + Laplacian Filtering (LF)  & 0.0290 & 0.0167 & 0.0761 & 0.0589 & 0.0329 & 0.0334 & 0.0455 & 0.1804 & \\
 & \quad + Stochastic Preconditioning (SP) + Laplacian Filtering (LF) & \cellcolor{best}0.02873 & 0.0167 & \cellcolor{best}0.0748 &  0.0599 &  0.0321 & \cellcolor{best}0.0335 & \cellcolor{best} 0.0450 & \cellcolor{best}0.1820 & \cellcolor{best}0.0591\\ \midrule

\multirow{3}{*}{MSE $\downarrow$}
 & Improved IMLS (Proposed Kernel) & 0.00123 & \cellcolor{best}0.00054 & 0.00546 & \cellcolor{best}0.00259 & \cellcolor{best}0.00128 & 0.00095 & 0.00204 & 0.02505 & 0.00489 \\
 & \quad + Stochastic Preconditioning (SP)  & 0.00172 & 0.000550 & 0.006465 & 0.002657 & 0.001428 & 0.000938 & 0.002019 &   0.025339 & \\
 & \quad + Laplacian Filtering (LF)  & 0.001199 & 0.000542 &  0.005344 & 0.002609 & 0.001381 & 0.00938 & 0.002054 & 0.0254 & \\
 & \quad + Stochastic Preconditioning (SP) + Laplacian Filtering (LF) & \cellcolor{best}0.001163 & 0.000558 & \cellcolor{best}0.00534 &  0.00265 & 0.00129 & \cellcolor{best}0.000916 & \cellcolor{best}0.001192 & \cellcolor{best}0.00202 & \cellcolor{best}0.00189 \\

\bottomrule
\end{tabular}%
}
\end{table*}

\begin{table}[htbp]
\centering
\caption{Dataset-specific hyperparameters for Stochastic Preconditioning ($\alpha$) 
and Laplacian Filtering ($\lambda$). For each dataset, we select the values that 
achieve the best performance based on PSNR, SSIM, LPIPS and MSE over a grid of candidate settings.}
\label{tab:hyperparams}
\begin{tabular}{lcccccccc}
\toprule
Params & Lego & Chair & Drums & Ficus & Hotdog & Mic & Materials & Ship \\
\midrule
$\alpha$ (SP) & 0.0035 & 0.0025 & 0.0035 & 0.0025 & 0.0035 & 0.0008 & 0.0015 & 0.0015 \\
$\lambda$ (LF) & 0.8 & 0.8 & 0.8 & 0.8 & 0.8 & 0.1 & 0.8 & 0.8 \\
\bottomrule
\end{tabular}
\end{table}

\begin{figure*}[t]
    \includegraphics[width=0.8\linewidth]{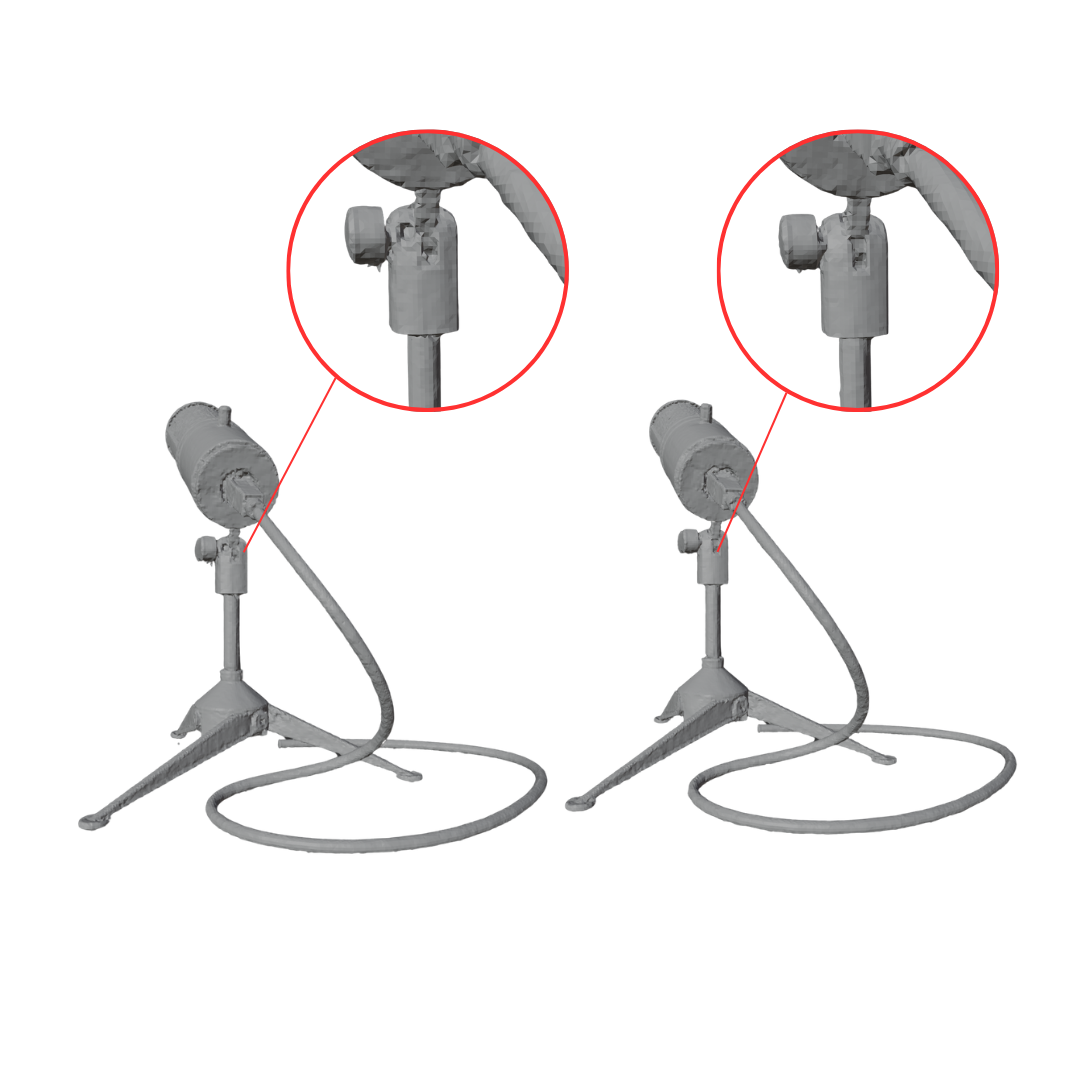}
    \includegraphics[width=0.8\linewidth]{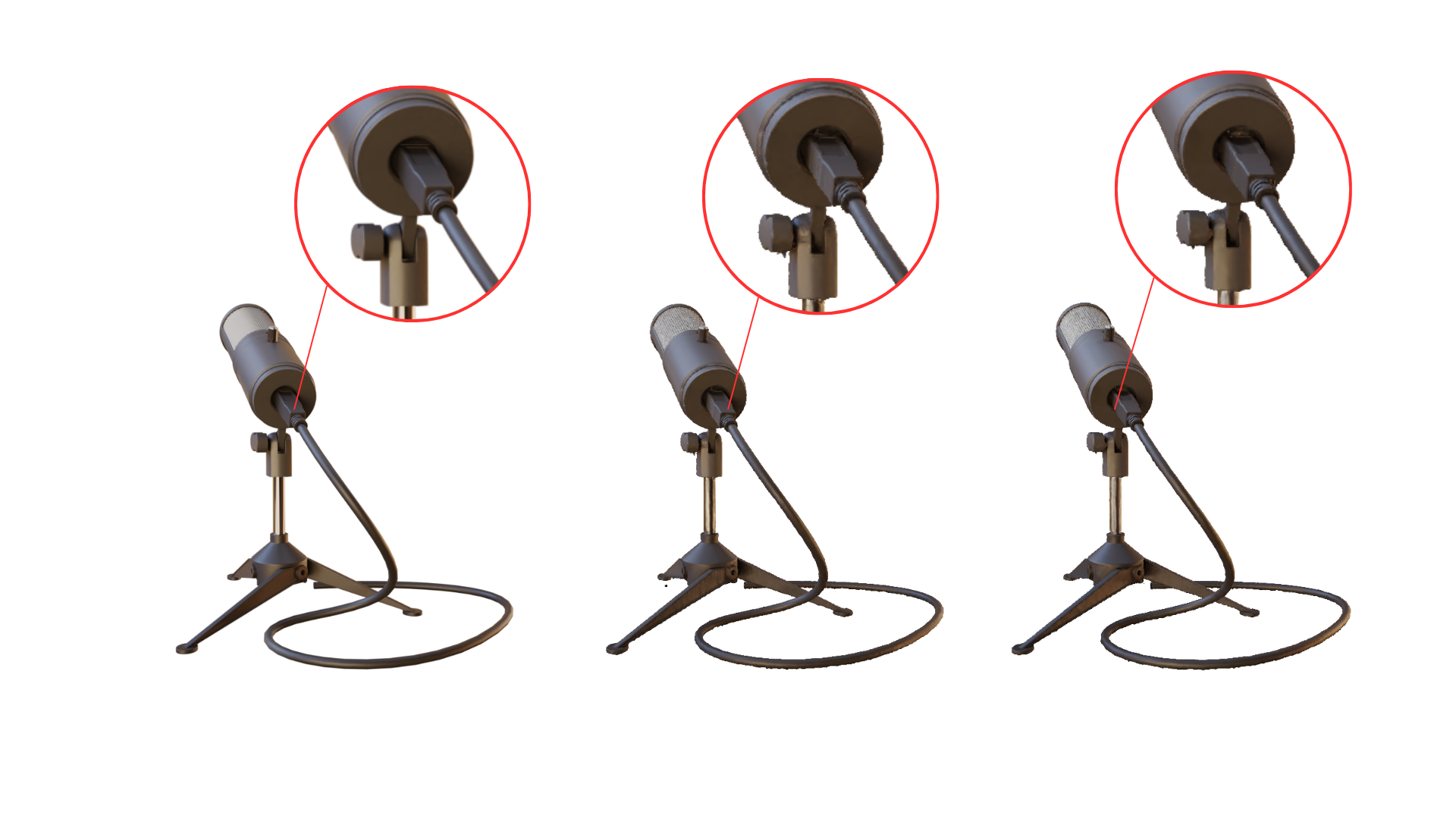}
    \caption{\textbf{Reconstructed Mesh comparison on the Mic scene (Top Row)}. Mesh reconstructions without \emph{(left)} and with \emph{(right)} SP and Laplacian, showing that the baseline exhibits floating artifacts and geometric inconsistencies, particularly in thin structures, while the full model produces smoother and more coherent surfaces.\\\\
    \textbf{Rendered images comparison on the Ficus scene (Bottom Row)}. \emph{(Left to Right)}.  Ground truth image, the reconstructed image without SP and Lap, and the reconstructed image with SP and Laplacian, along with corresponding zoomed-in regions. The full model (right) reduces artifacts and improves geometric fidelity, better matching the ground truth.}
    \label{fig:ablation_sp_lap_mic}
\end{figure*}

% {Figures/ficus_comp.png}
%     \includegraphics[width=0.5\linewidth]{Figures/lego.png}
%     \includegraphics[width=0.5\linewidth]{Figures/lego_better.png}
%     \includegraphics[width=0.5\linewidth]{Figures/lego_1.png}
%     \includegraphics[width=0.5\linewidth]{Figures/lego_better1.png}

\begin{figure}[!htbp] 
\centering
        \includegraphics[width=\linewidth]{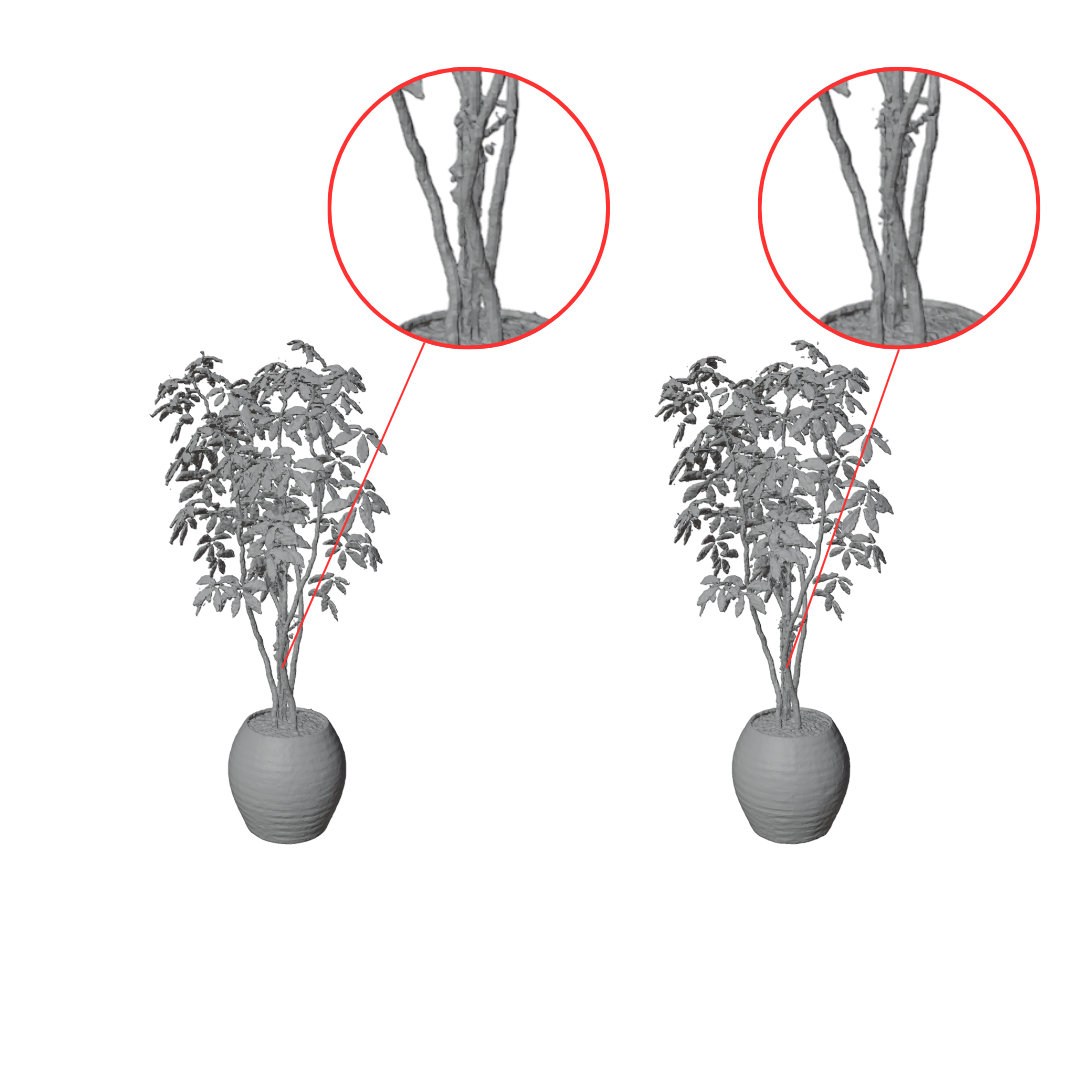}
        \vspace{-1.5cm}
        \includegraphics[width=\linewidth]{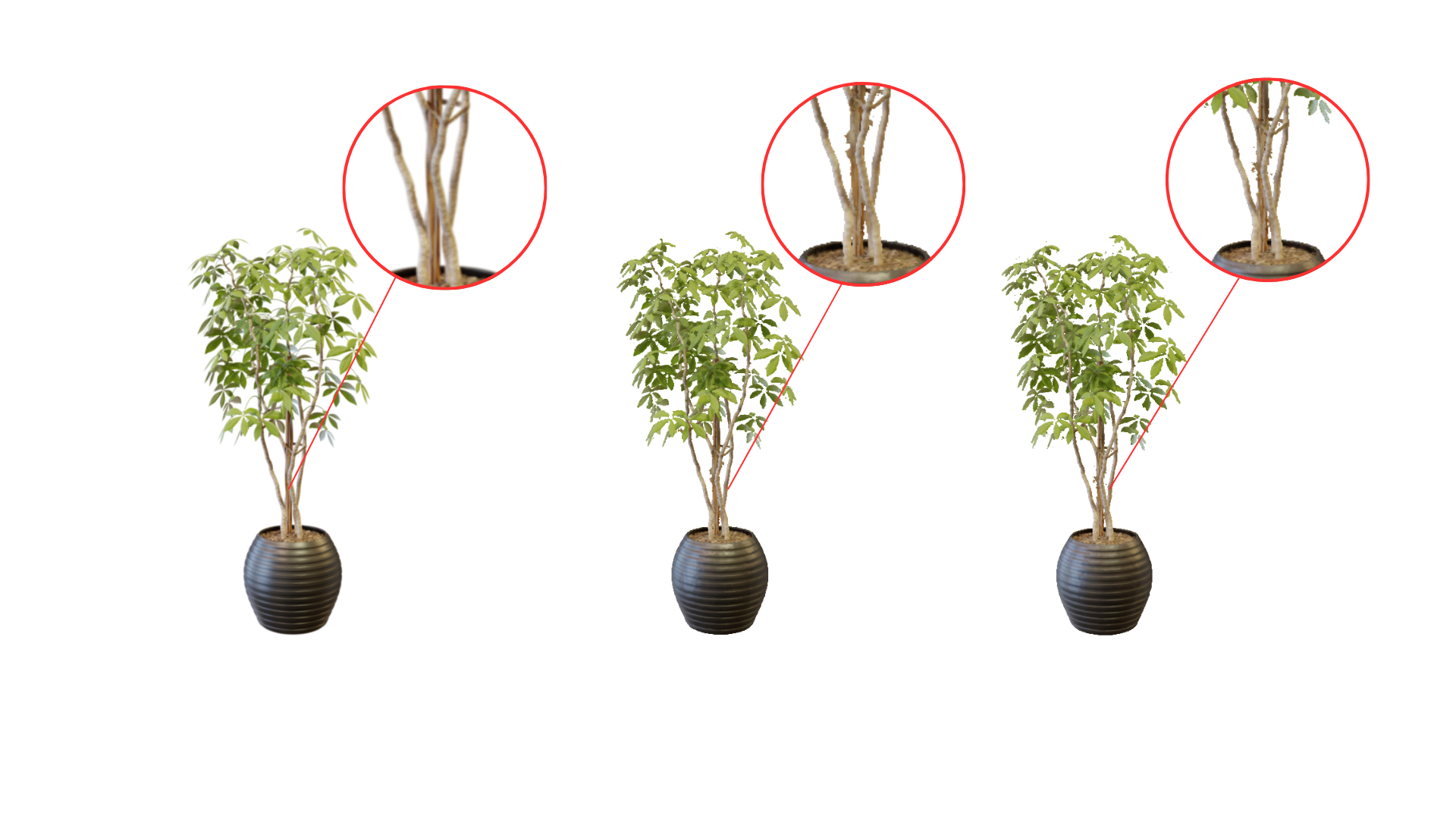}
        \caption{\textbf{Reconstructed Mesh comparison on the Ficus scene (Top Row)}. \emph{(Left)} Reconstruction without stochastic preconditioning (SP) and Laplacian regularization exhibits floating artifacts and structural inconsistencies. \emph{(Right)} Incorporating SP and Laplacian produces smoother and more coherent surfaces.\\\\
        \textbf{Rendered images comparison on the Ficus scene (Bottom Row)}. \emph{(Left to Right)}. Ground truth, reconstruction without SP and Lap, and reconstruction with SP and Lap. The full model (right) reduces artifacts and better preserves fine geometric details.}
    \label{fig:ablation_ficus}
\end{figure}

\begin{figure}[!htb]
        \includegraphics[width=0.5\linewidth]{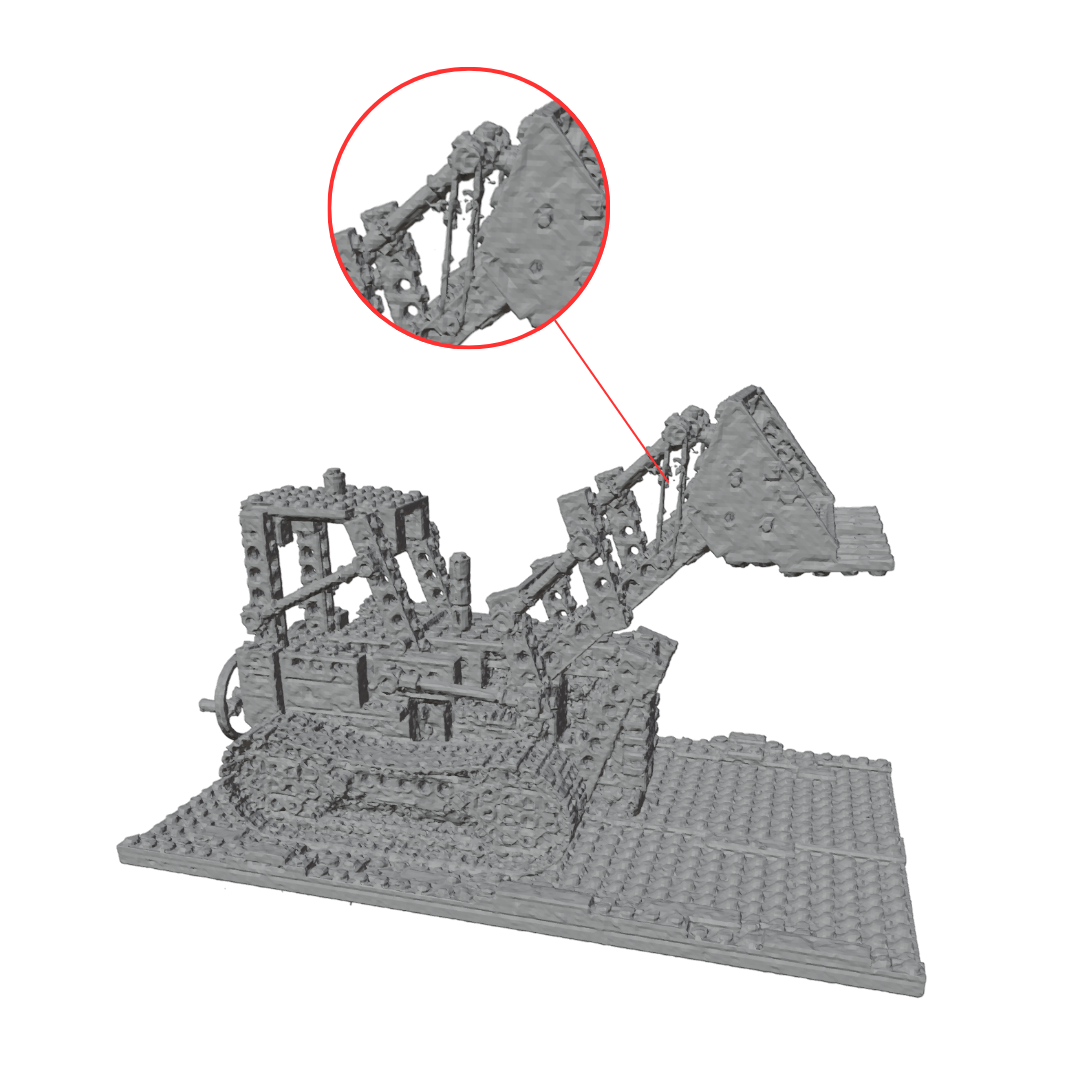}~%
        %\caption{Mesh reconstruction without SP and Lap.}
        \includegraphics[width=0.5\linewidth]{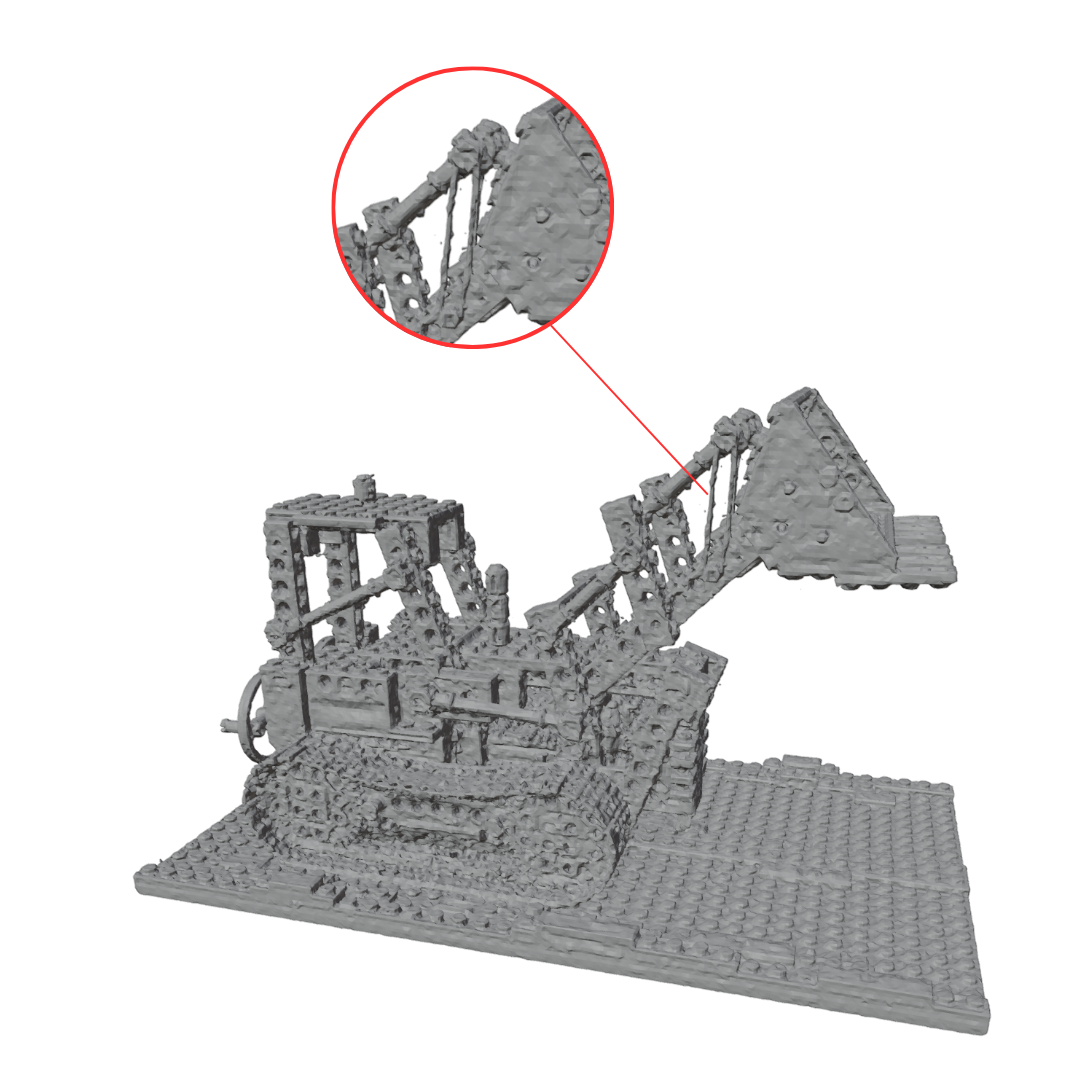}
        %\caption{Mesh reconstruction with SP and Lap.}
        \includegraphics[width=\linewidth]{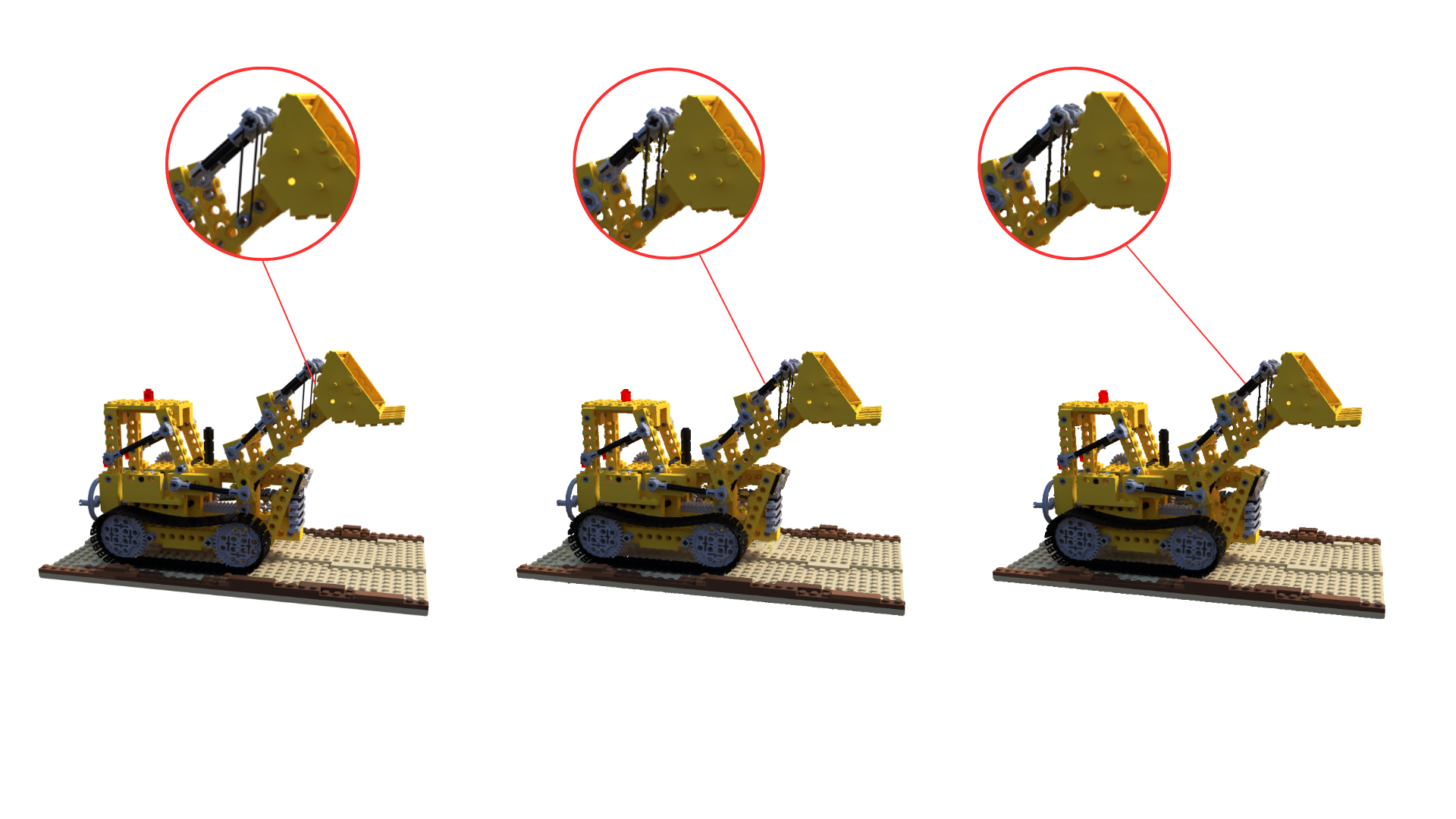}
        \caption{\textbf{Reconstructed Mesh comparison on the Lego scene (Top Row)}. \emph{(Left)} Reconstruction without stochastic preconditioning (SP) and Laplacian regularization exhibits floating artifacts and structural inconsistencies. \emph{(Right)} Incorporating SP and Laplacian produces smoother and more coherent surfaces.\\\\ 
        \textbf{Rendered images comparison on the Lego scene (Bottom Row)}. \emph{(Left to Right)} ground truth, reconstruction without SP and Laplacian, and reconstruction with SP and Laplacian. The full model improves geometric fidelity and reduces artifacts.}
    \label{fig:ablation_lego}  
\end{figure}

\begin{figure}[h!tb]
        \includegraphics[width=0.5\linewidth]{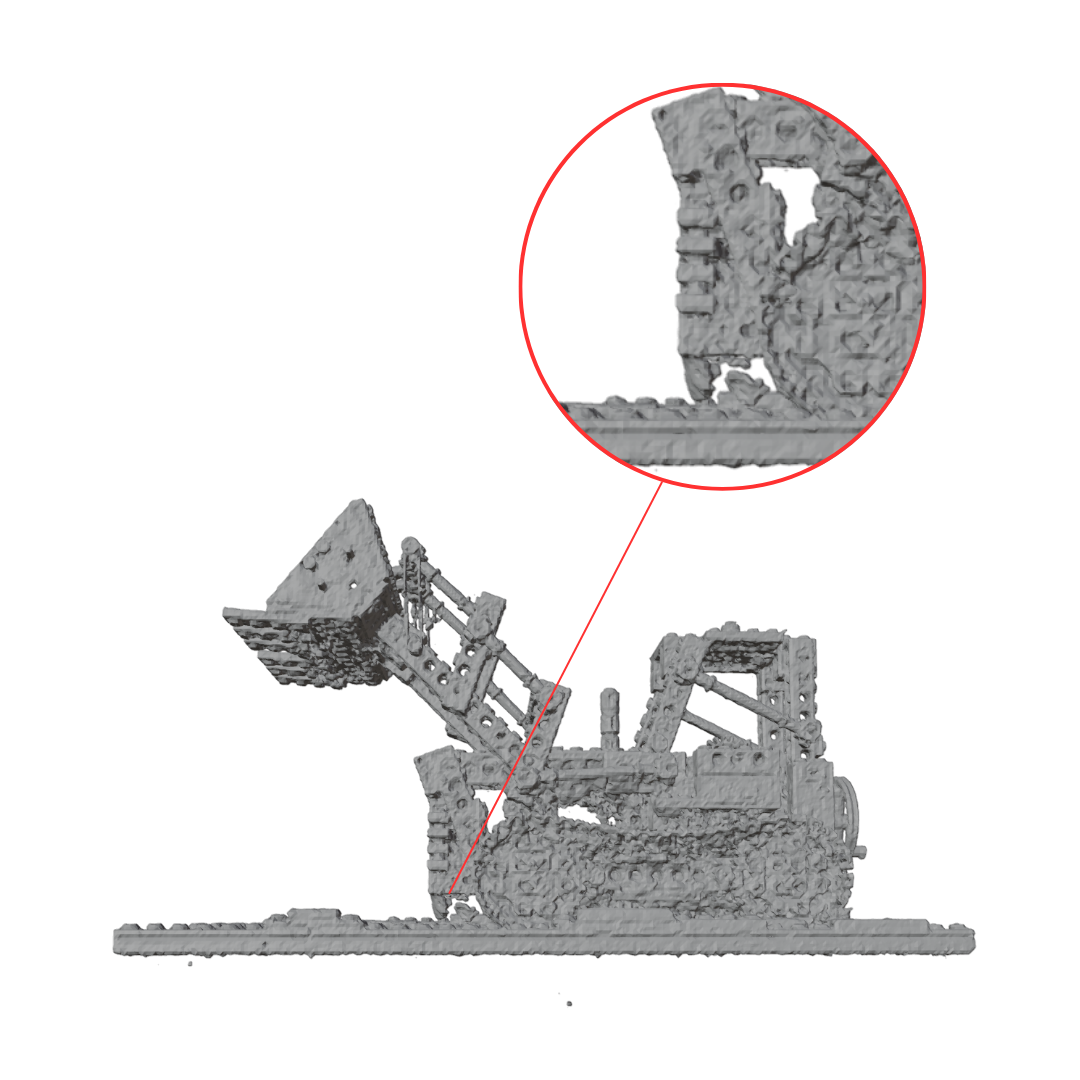}~%
        \includegraphics[width=0.5\linewidth]{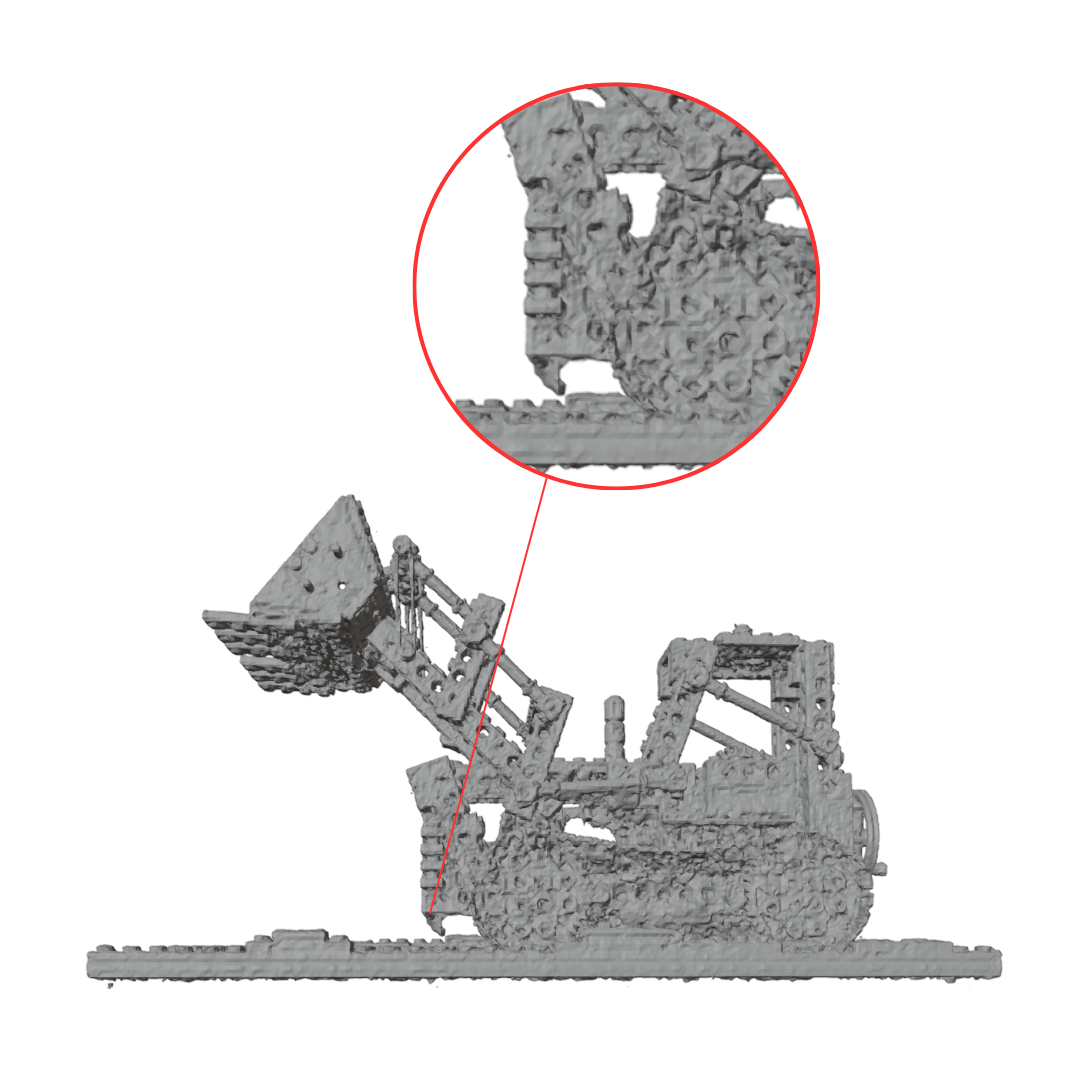}
        \includegraphics[width=\linewidth]{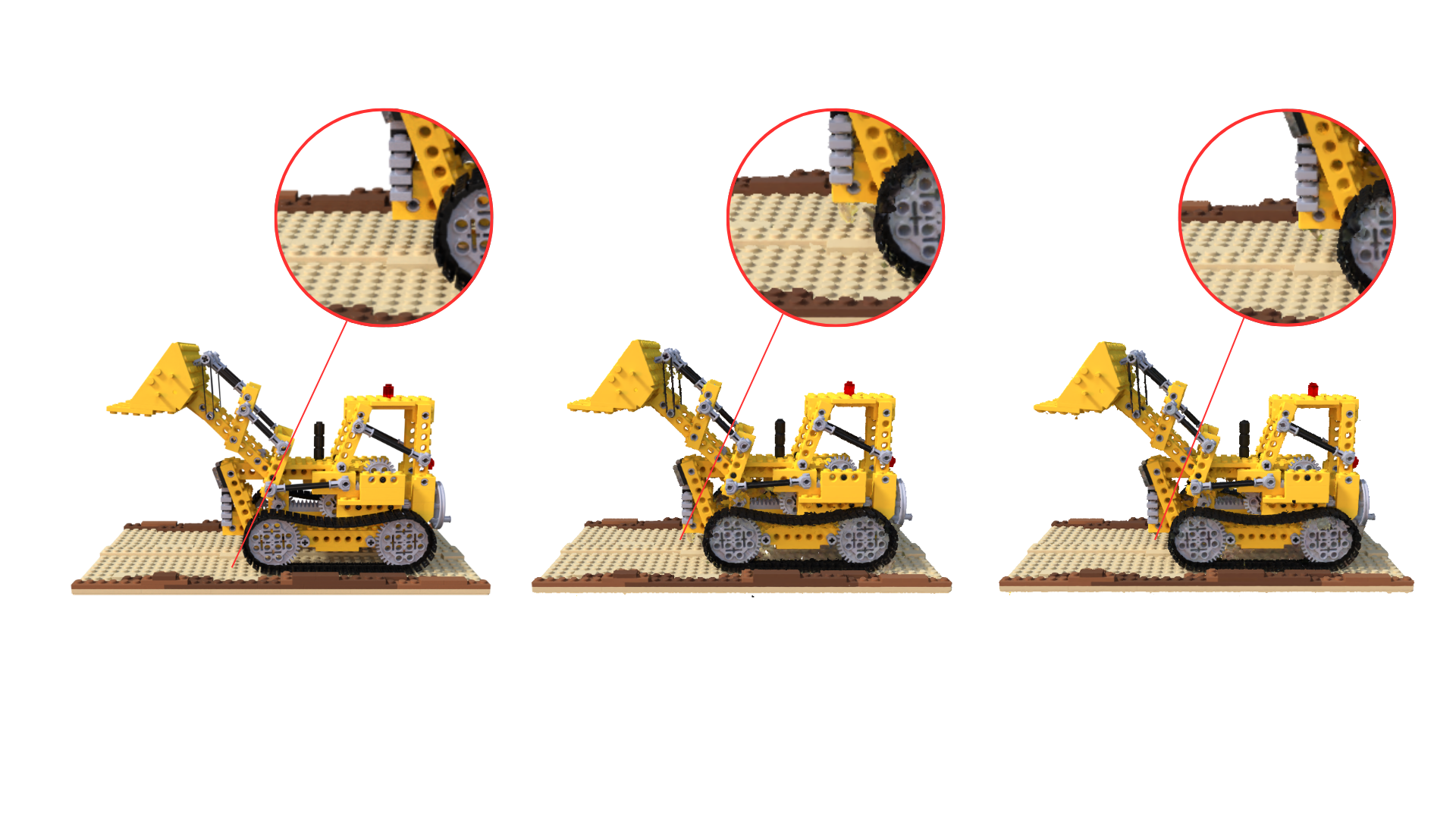}
        \caption{\textbf{Reconstructed Mesh comparison on the Lego scene (Top Row)} (View from a different angle). \emph{(Left)} Reconstruction without stochastic preconditioning (SP) and Laplacian regularization exhibits floating artifacts and structural inconsistencies. \emph{(Right)} Incorporating SP and Laplacian produces smoother and more coherent surfaces.\\\\ 
        \textbf{Rendered images comparison on the Lego scene (Bottom Row)}. \emph{(Left to Right)}. Ground truth, reconstruction without SP and Laplacian, and reconstruction with SP and Laplacian. The full model (right) improves geometric fidelity and reduces artifacts.}
        \label{fig:ablation_lego_diff}
\end{figure}

\begin{figure}[!htb]
        \includegraphics[width=\linewidth]{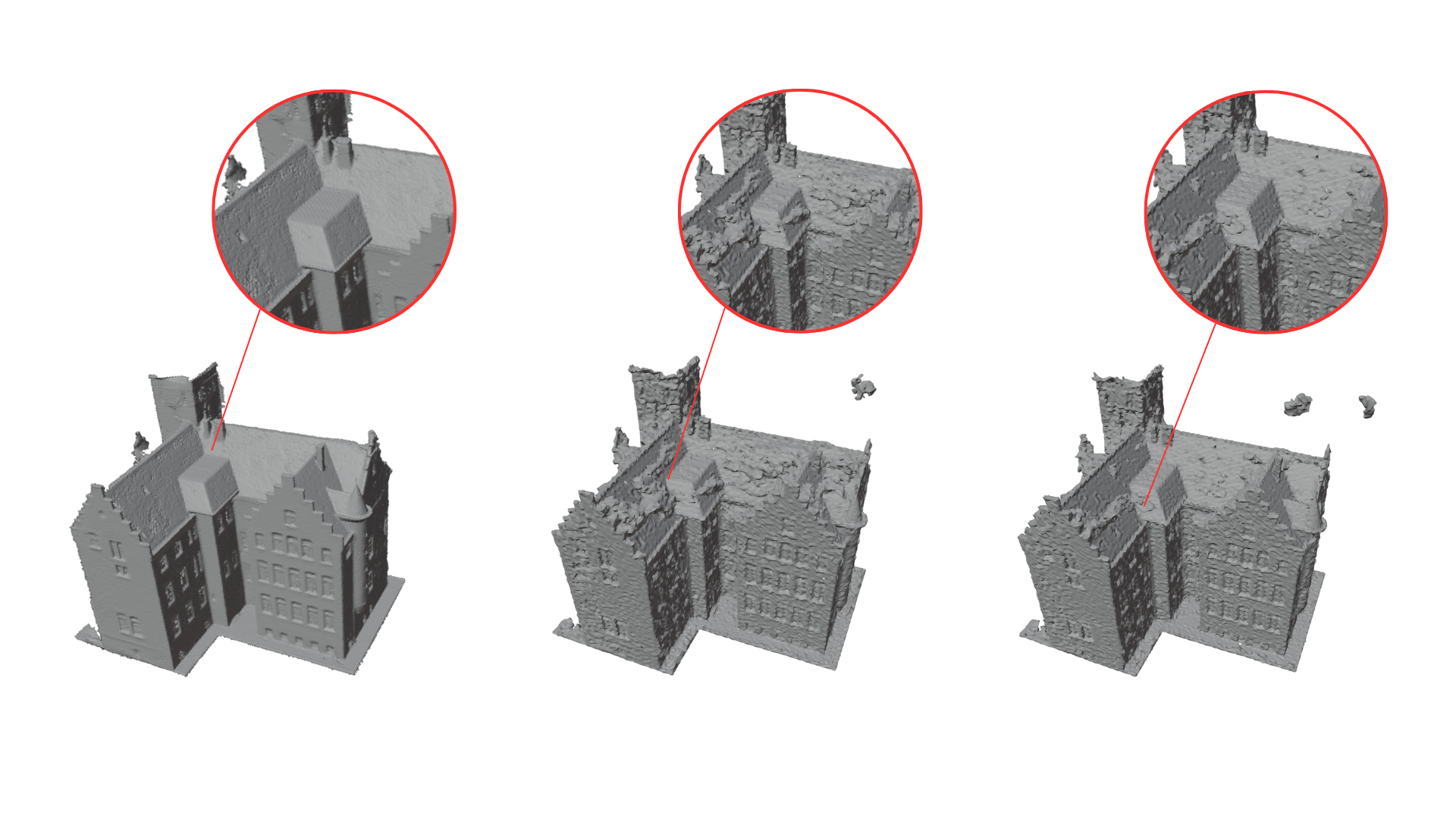}
        \includegraphics[width=\linewidth]{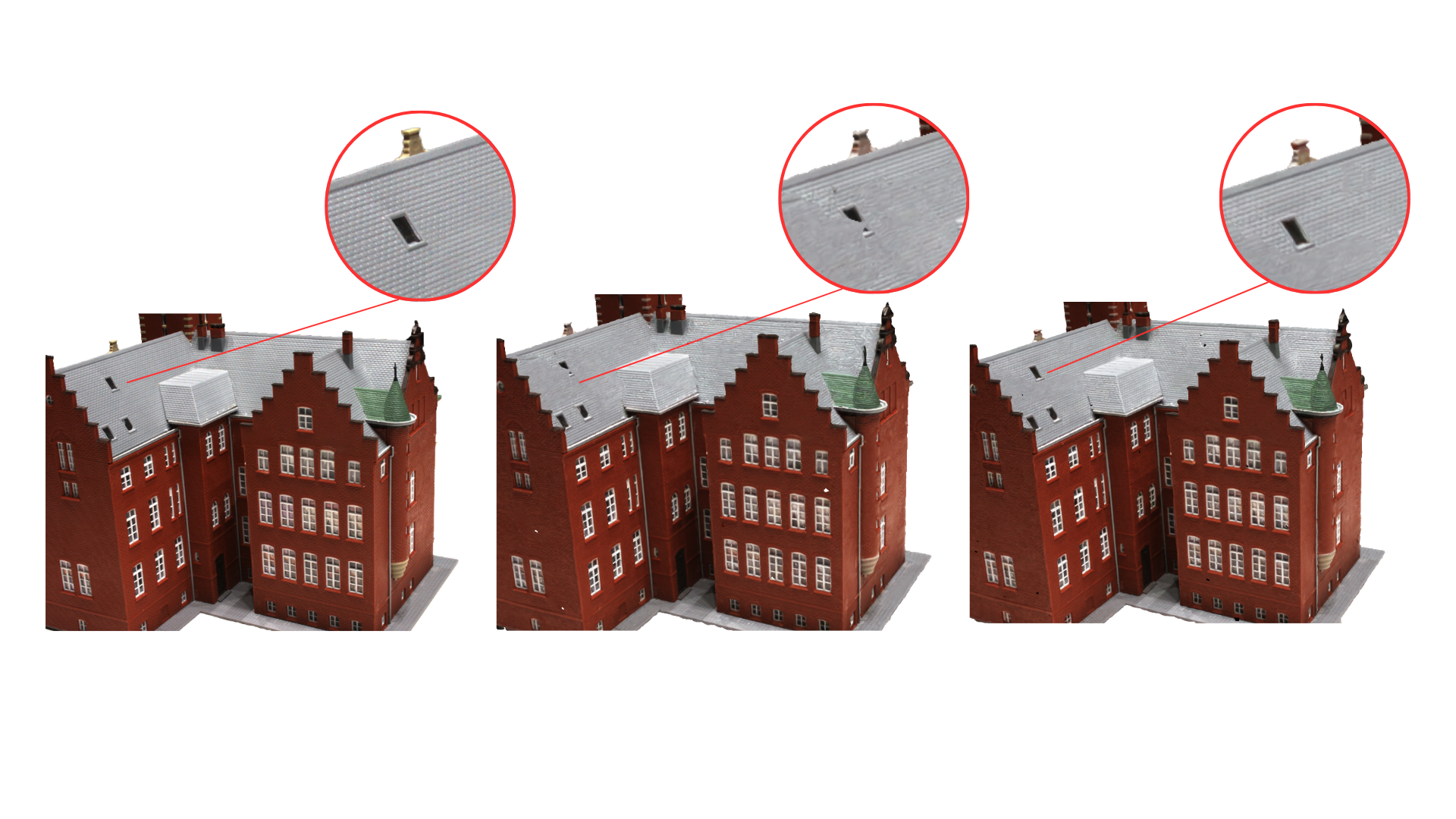}
        \caption{\textbf{Reconstructed Mesh comparison on the scan24 scene (Top Row)}. \emph{(Left to Right)}. Ground truth, reconstruction without stochastic preconditioning (SP), and Laplacian regularization, and reconstruction with SP and Laplacian. Reconstruction without stochastic preconditioning and Laplacian regularization exhibits floating artifacts and structural inconsistencies. Reconstruction with SP and Laplacian mitigates them.\\\\ 
        \textbf{Rendered images comparison on the scan24 scene (Bottom Row)}. \emph{(Left to Right)}. Ground truth, reconstruction without SP and Laplacian, and reconstruction with SP and Laplacian. The full model (right) improves geometric fidelity and reduces artifacts.}
        \label{fig:ablation_scan24_diff}
\end{figure}

\begin{figure}[!htb]
        \includegraphics[width=\linewidth]{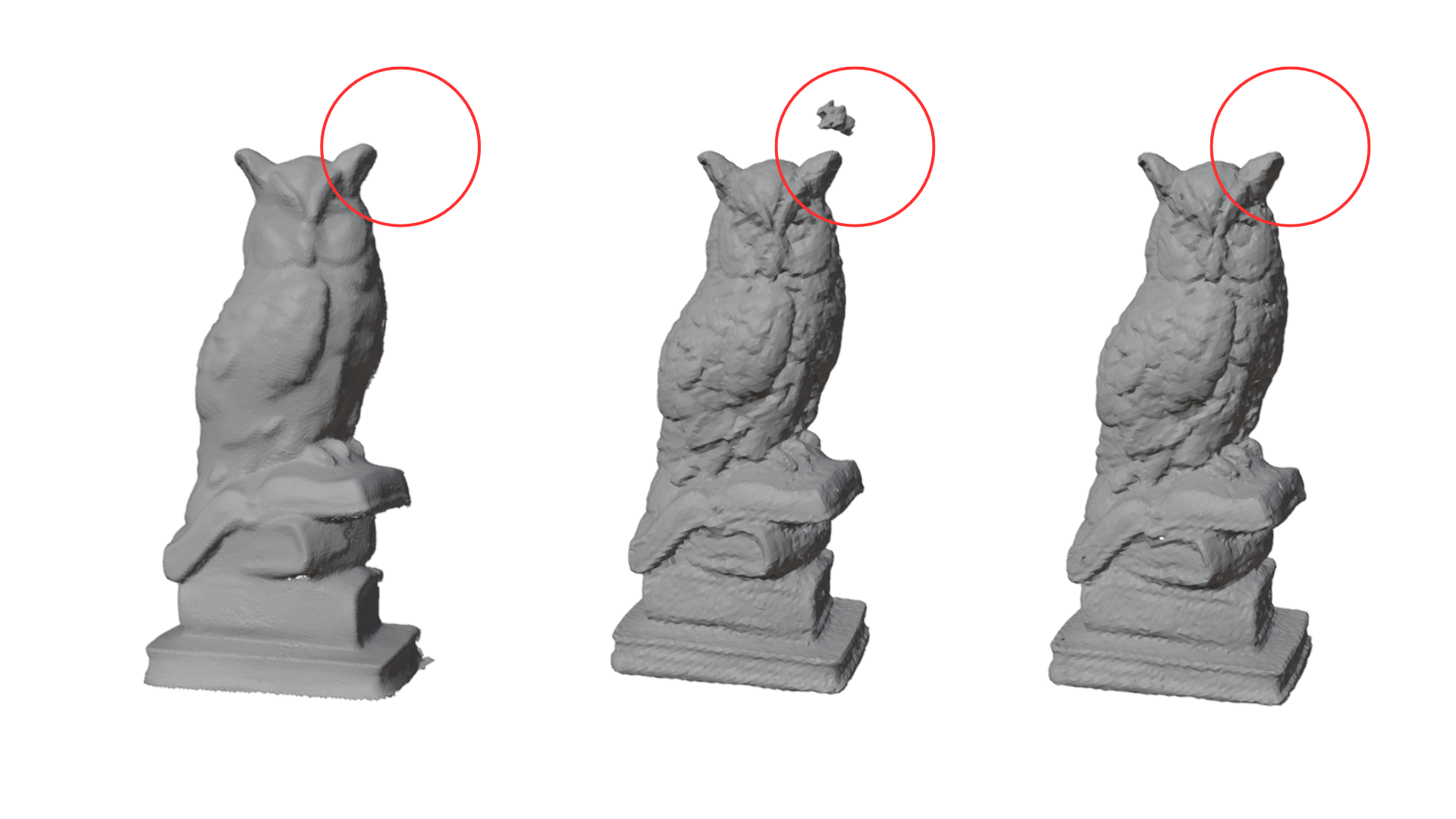}
        \caption{\textbf{Reconstructed Mesh comparison on the scan122 scene}. \emph{(Left  to Right)}. Ground truth, reconstruction without stochastic preconditioning (SP), and Laplacian regularization, reconstruction with stochastic preconditioning (SP), and Laplacian regularization. Reconstruction without stochastic preconditioning and Laplacian regularization exhibits floating artifacts and structural inconsistencies. Reconstruction with SP and Laplacian removes them.\\\\}
        \label{fig:ablation_scan122_diff}
\end{figure}

As shown in Fig.~\ref{fig:ablation_sp_lap_mic},~\ref{fig:ablation_ficus},~\ref{fig:ablation_lego},~\ref{fig:ablation_lego_diff}, ~\ref{fig:ablation_scan24_diff},  ~\ref{fig:ablation_scan122_diff}, the absence of stochastic preconditioning and Laplacian filtering leads to noticeable high-frequency artifacts and fragmented geometry. These issues are especially prominent in regions with fine structures and high curvature. By introducing stochastic preconditioning and Laplacian regularization, the reconstruction becomes significantly smoother and structurally consistent, with reduced noise and improved surface continuity. As shown in Table~\ref{tab:ablation}, the PSNR, SSIM, LPIPS, and MSE do not consistently capture the improvement from adding SP and the Laplacian. However, the visual quality is much better, as evident from the figures. Table~\ref{tab:hyperparams} lists the parameters that produce the best output for each scene. 
%\vspace{-0.75cm}

%%%%%%%%%%%%%%%%%%%%%%%%%%%%%%%%%%%%%%%%%%%%%%%%%%%%%%%%%%%%

%\newpage
%\clearpage
%\input{checklist}

% \section{To ADD}

\end{document}